\documentclass[final]{cvpr}

\usepackage{times}
\usepackage{epsfig}
\usepackage{graphicx}
\usepackage{amsmath}
\usepackage{amssymb}
\usepackage{bbm}
\usepackage{booktabs}
\usepackage{makecell}
\usepackage{multirow}
\usepackage{cuted}
\usepackage[font={small}]{caption}

\usepackage[pagebackref=true,breaklinks=true,colorlinks,bookmarks=false]{hyperref}

\def\cvprPaperID{0819} 
\def\confYear{CVPR 2021}

\begin{document}

\title{Everything's Talkin': Pareidolia Face Reenactment}

\author{Linsen Song\textsuperscript{1,2\textasteriskcentered}\quad Wayne Wu\textsuperscript{3,4}\thanks{Equal Contribution}\quad Chaoyou Fu\textsuperscript{1,2}\quad Chen Qian\textsuperscript{3}\quad Chen Change Loy\textsuperscript{4}\quad Ran He\textsuperscript{1,2}\thanks{Corresponding Author}\\
\textsuperscript{1}NLPR \& CRIPAC, CASIA\quad \textsuperscript{2}University of Chinese Academy of Sciences\\
\textsuperscript{3}SenseTime Research\quad \textsuperscript{4}Nanyang Technological University\\
{\tt\small songlinsen2018@ia.ac.cn, \{wuwenyan,qianchen\}@sensetime.com,}\\
{\tt\small \{chaoyou.fu,rhe\}@nlpr.ia.ac.cn, ccloy@ntu.edu.sg}
\vspace{-1.8cm}
}

\maketitle

\begin{strip}
    \centering
    \includegraphics[width=0.9\textwidth]{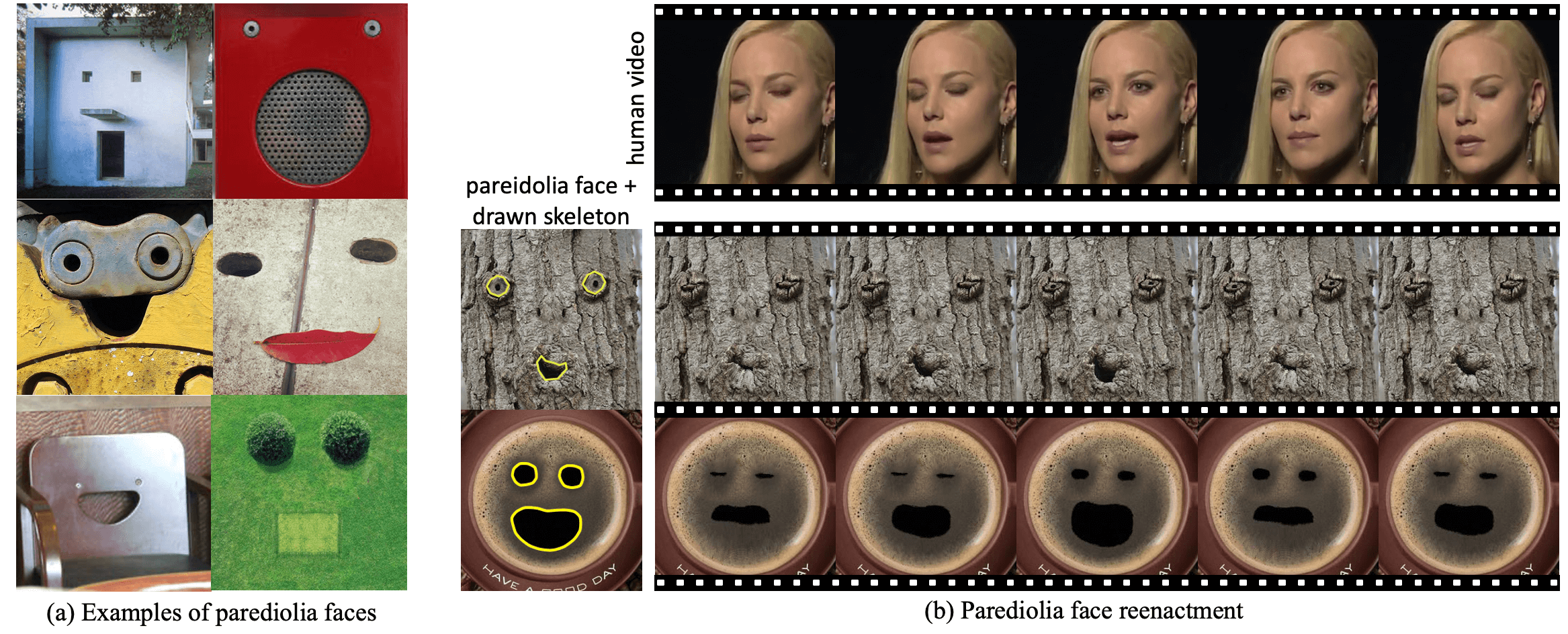}
    \vspace{-0.2cm}
    \captionof{figure}{\textbf{Pareidolia Face Reenactment.}
    We propose an unsupervised method for static illusory faces to become animated by reenacting the human face in a video. The eyes and mouth of the illusory faces, as defined by users, move in tandem with those of the human in the video, simultaneously.
    }
    \label{fig:first}
\end{strip}

\begin{abstract}
We present a new application direction named Pareidolia Face Reenactment, which is defined as animating a static illusory face to move in tandem with a human face in the video. For the large differences between pareidolia face reenactment and traditional human face reenactment, two main challenges are introduced, i.e., shape variance and texture variance. In this work, we propose a novel Parametric Unsupervised Reenactment Algorithm to tackle these two challenges. Specifically, we propose to decompose the reenactment into three catenate processes: shape modeling, motion transfer and texture synthesis. With the decomposition, we introduce three crucial components, i.e., Parametric Shape Modeling, Expansionary Motion Transfer and Unsupervised Texture Synthesizer, to overcome the problems brought by the remarkably variances on pareidolia faces. Extensive experiments show the superior performance of our method both qualitatively and quantitatively. Code, model and data are available on our project page\footnote{ \url{https://wywu.github.io/projects/ETT/ETT.html}}.
\end{abstract}

\vspace{-0.8cm}
\section{Introduction}

\textit{It's not often that you look at your meal to find it staring back at you. But when Diane Duyser picked up her cheese toastie, she was in for a shock. “I went to take a bite out of it, and then I saw this lady looking back at me,” she told the Chicago Tribune. “It scared me at first.”}~\cite{news}.

The phenomenon described in this BBC news is called face pareidolia, a natural inclination of the human brain to perceive illusory faces that do not actually exist~\cite{hao2018face,wardle2020rapid}. In this work, we attempt to bring this interesting imagination into reality by animating. As shown in Fig.~\ref{fig:first} (b), we propose a new application direction named \textit{``Pareidolia Face'' Reenactment}, which is defined as animating illusory faces by the motion extracted from human faces automatically.

Pareidolia face reenactment, has large potential usages in filmmaking~\cite{thies2016face2face,kim2019neural}, cartoon production~\cite{yi2020animating,zhou2020makeittalk} and mixed reality~\cite{thies2018facevr,zhang2020facial}, which always requires a massive labor of professional animators. Mostly related, face reenactment~\cite{garrido2014automatic,thies2016face2face,kim2018deep,wu2018reenactgan} is becoming an emerging topic in recent years. However, all of these methods are designed specifically for human faces, of which rich priors like facial landmarks~\cite{zakharov2019few,ha2020marionette} or 3D face models~\cite{thies2016face2face,kim2019neural} can be utilized. But, all of these priors are unachievable for pareidolia faces. Moreover, large-scale face datasets~\cite{bulat2017far,sagonas2013300} with massive annotations are sufficient for human faces, which are also unreachable for pareidolia faces. Reenacting pareidolia faces by human portrait videos is still an open question.

The main challenges for pareidolia face reenactment can be summarized into two large \textit{variances}, \ie, shape variance and texture variance. \textit{Shape variance} means that the boundary shapes of facial parts are remarkably diverse, such as circular, square and moon-shape mouths as shown in Fig.~\ref{fig:first} (a). For human faces, landmarks are always used as the intermediary to transfer motions~\cite{chen2019hierarchical,ha2020marionette,zhou2020makeittalk}. However, landmark suffers from the tightly coupling with the shape/size of facial parts. It cannot be used as the intermediary to perform a precise motion transfer from the source human face to the target pareidolia face. The shapes of target faces will be affected by the source ones' easily. Also, it is difficult to define the meaning of landmarks' annotation for complex shapes, \eg, the tree's mouth in Fig.~\ref{fig:first} (b). Thus, it is challenging to design a universal shape representation to transfer motion from human faces to pareidolia faces.

\textit{Texture variance} means the textures of pareidolia faces are remarkably diverse, such as wood, downy and metal textures as shown in Fig.~\ref{fig:first} (a). Also, the texture distribution of pareidolia faces is extremely discrete, since there is even no two faces with a similar texture. For human face, previous works always deployed 3D facial models~\cite{thies2016face2face,kim2019neural} or GAN-based generator~\cite{ha2020marionette,zakharov2019few} in texture synthesis. However, there is no 3D face model that can be leveraged to model pareidolia face. Also, for the GAN-based synthesis, large-scale labeled datasets with landmark-image pairs are always needed to train a generator~\cite{nirkin2019fsgan,wu2018reenactgan}. But, there exists no dataset or annotation for pareidolia faces, which makes the strong supervision with paired data out of action for texture synthesis. Thus, synthesizing the textures of pareidolia faces is challenging, without a 3D model or annotated data.

In this work, we propose a novel Parametric Unsupervised Reenactment Algorithm, to tackle the pareidolia face reenactment problem. First, to solve the shape variance challenge, we propose a \textit{Parametric Shape Modeling} technique, in which we introduce B\'{e}zier Curve~\cite{farin2002handbook}, a classic parametric technology in computer graphics, to represent the boundary shapes of facial parts of both the source and target faces with a set of control points. With the parametric modeling of boundaries on target pareidolia face, control points of the B\'{e}zier Curve can locally modify the curve while keeping its global structure unchanged, even with large shape variance.

With the robust shape representation, a na\"{i}ve solution to transfer motion is to directly adapt the human face' control points to the pareidolia face. However, the transferred motion so far only decides the movement of the facial boundaries of the pareidolia face, which is a local motion and cannot be used to drive the whole face. Thus, we propose an \textit{Expansionary Motion Transfer} technique to get a global motion representation named motion field for a natural animation, in which a Motion Spread strategy is designed to propagate the transferred motion from boundary to the whole face and a First-order Motion Approximation strategy is designed to refine the motion field further.

While the motion has been successfully transferred, the next step is to use the motion field to deduce an image with high-quality textures. Reviewing the challenge of texture variance, we propose an \textit{Unsupervised Texture Synthesizer} to address it in an AutoEncoder framework with a carefully designed Feature Deforming Layer. High-quality textures can be synthesized successfully, while neither 3D model, nor large-scale face datasets with annotations are needed.

We summarize our contributions as follows: 1) We make the first attempt to animate pareidolia faces by the facial motion derived from the human faces. 2) We propose a novel Parametric Unsupervised Reenactment Algorithm to tackle pareidolia face reenactment, with three crucial components, \ie, Parametric Shape Modeling, Expansionary Motion Transfer and Unsupervised Texture Synthesizer. 3) Extensive experiments present the superior performance of our method and the effectiveness of each component.

\begin{figure*}[t!]
    \centering
    \includegraphics[width=0.9\linewidth]{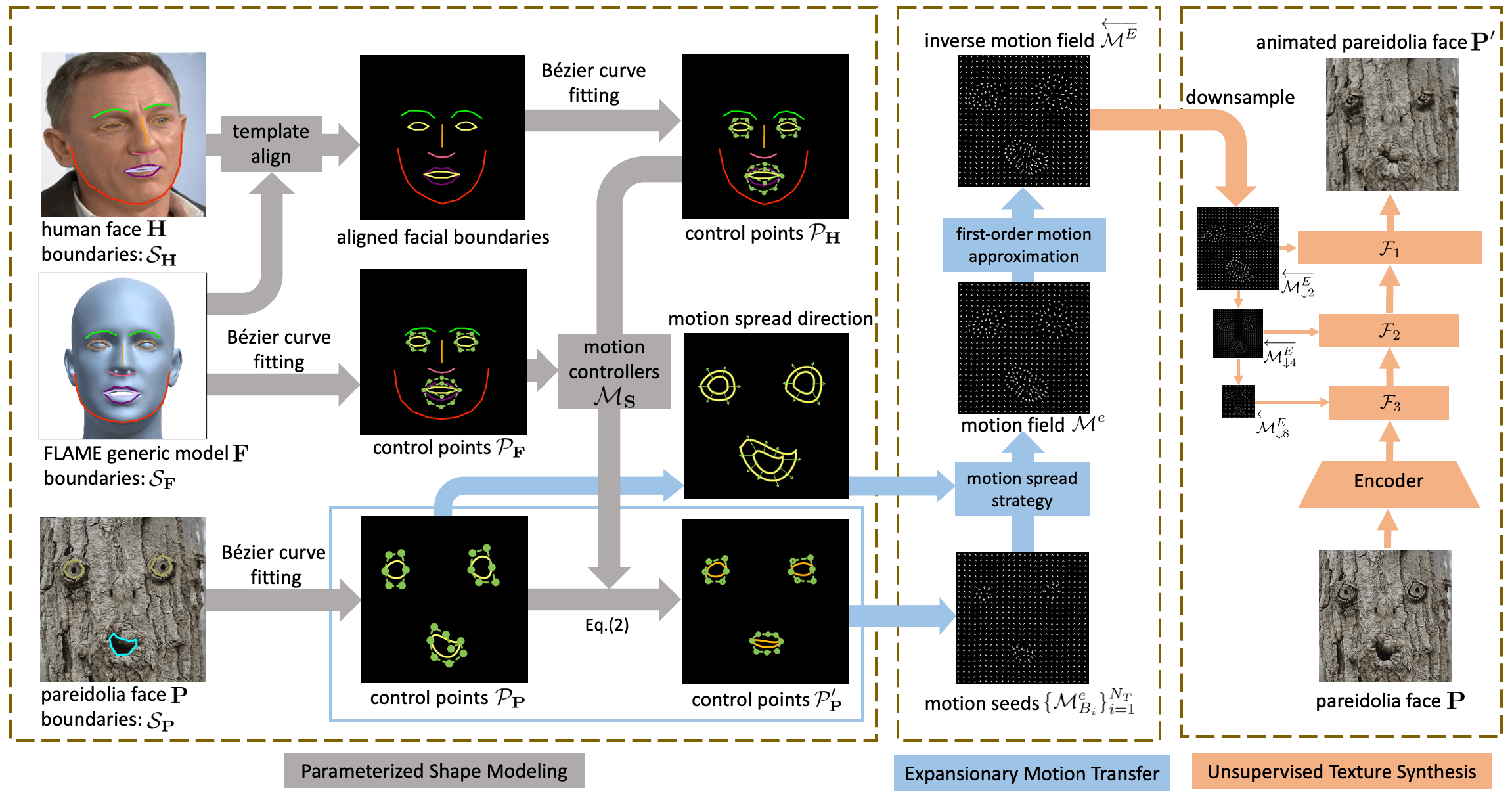}
    \caption{\textbf{Parametric Unsupervised Reenactment Algorithm.} We separate the proposed algorithm into three components: (a) Parametric Shape Modeling. First, we model the facial boundary of three kinds of faces on the left-most column, with B\'{e}zier Curve fitting. Then the facial boundaries can be represented as a set of control points. By simply adapting the motion controller inferred from the human and FLAME face, we can get the animated control points of the pareidolia face $\mathbf{P}'$. (b) Expansionary Motion Transfer: The animated control points are then converted into an optical flow map as motion seeds, which represent a local motion of the target pareidolia face. Motion Spread and First-order Motion Approximation are proposed to extend the local motion seed to a global motion field $\mathcal{M}^{e}$. (c) Unsupervised Texture Synthesizer: With the motion field and the raw pareidolia face as conditions, we can synthesize the final animated pareidolia face.
    }
    \label{fig:arch}
    \vspace{-2mm}
\end{figure*}

\section{Related Work}

\subsection{Face Reenactment}
Face reenactment refers to transferring motion patterns from one face to another one, including both graphics-based~\cite{thies2015real,averbuch2017bringing} and learning-based~\cite{ha2020marionette,huang2020learning,nirkin2019fsgan,suwajanakorn2017synthesizing} methods.
The former mainly relies on 3DMMs~\cite{blanz1999morphable}.
Benefiting from the face fitting capacity of 3DMMs, recent methods, e.g. Face2Face~\cite{thies2016face2face} and DVP~\cite{kim2018deep}, can reenact a given face by adjusting the fitted parameters.
Nevertheless, 3DMMs, designed for human faces, are inapplicable to our pareidolia faces.
The latter mainly resorts to Deep Neural Networks (DNNs).
Thanks to the powerful expression capacity of DNNs, GAN~\cite{goodfellow2014generative}-based methods~\cite{hu2018pose,song2019geometry,fu2021high,fu2020dvg,fu2019dual,kaisiyuan2020mead} like ReenactGAN~\cite{wu2018reenactgan} can achieve face reenactment via learning a mapping from a source face to a target one. 
However, these methods usually require a large amount of paired training data, which are unavailable in our task.
Furthermore, apart from reenacting human faces, there are also some methods try to animate non-human faces, such as the cat face~\cite{wayne2019transgaga,qian2019make} or the cartoon face~\cite{zhou2020makeittalk}.
Recent audio-driving method~\cite{zhou2020makeittalk} animates cartoon faces while their labeled 68 facial landmarks correspond to human faces.
These methods are all driven by facial landmarks, which are inapplicable in our pareidolia face reenactment task.

\subsection{Geometric Shape Modeling}
There are two means to model facial geometric shape: implicit or explicit modeling.
The former directly disentangles shape representations from faces via elaborately designed networks and training manners~\cite{tran2017disentangled,li2019faceshifter,burkov2020neural}.
The latter leverages extra auxiliary models to present facial shape information, \textit{e.g.} facial landmarks~\cite{hu2018pose,zakharov2019few,geng2018warp} or 3D parameters~\cite{thies2020neural,song2020everybody}.
B\'{e}zier curves are a very popular tool in computer-aided design~\cite{glassner2013graphics}, computer graphics and interactive curve design~\cite{salomon2007curves}.
Recently, B\'{e}zier curves are incorporated with deep learning methods like CNNs~\cite{krizhevsky2017imagenet} and GANs~\cite{goodfellow2014generative} in tasks includes parametric skeleton extraction~\cite{liu2019parametric} and sketch generation from human drawing~\cite{song2020beziersketch}.
In this paper, we explore the application of B\'{e}zier curves in modeling the boundary of facial parts.

\section{Methodology}
The architecture of our proposed method is shown in Fig.~\ref{fig:arch}, which is separated into three main components: Parametric Shape Modeling, Expansionary Motion Transfer and Unsupervised Texture Synthesizer.
First, we extract the boundaries of both human and pareidolia faces. We build a robust shape model for the facial boundaries based on the \textit{B\'{e}zier Curve} and represent the motion as the \textit{Motion Controller} (Sec.~\ref{sec:shape_motion}).
Then, we get an optical flow map named motion seed to represent the transferred local motion at the facial boundaries.
In order to animate the whole pareidolia face, we propose \textit{Motion Spread} and \textit{First-order Motion Approximation} strategy to induce a global motion representation named motion field. (Sec.~\ref{sec:motion_transfer_densify}).
At last, we propose an unsupervised network with a carefully designed \textit{Feature Deforming Layer} to synthesize high-quality animated texture conditioned on the static pareidolia face and the motion field. (Sec.~\ref{sec:texture_syn})

\subsection{Parametric Shape Modeling}
\label{sec:shape_motion}
Reviewing the remarkable shape variance exists in facial parts' shapes of human and pareidolia faces. B\'{e}zier Curve can be edited locally while remain the whole structure. Thus, we introduce it to robustly model the shapes of facial parts. In this section, we first describe the Shape Modeling of facial boundaries in detail, then introduce Motion Controller, the motion representation based on our shape modeling fashion.

\noindent
\textbf{Shape Modeling by B\'{e}zier Curves.}
Composite B\'{e}zier Curve~\cite{shikin1995handbook}, defined as a piecewise B\'{e}zier Curve~\cite{farin2002handbook}, is exploited to fit the facial boundaries since composite B\'{e}zier Curves can freely model complex boundary and each of its control points can regulate the curve locally and do not break the curve's global structure.

\begin{figure}[t]
    \centering
    \includegraphics[width=0.95\linewidth]{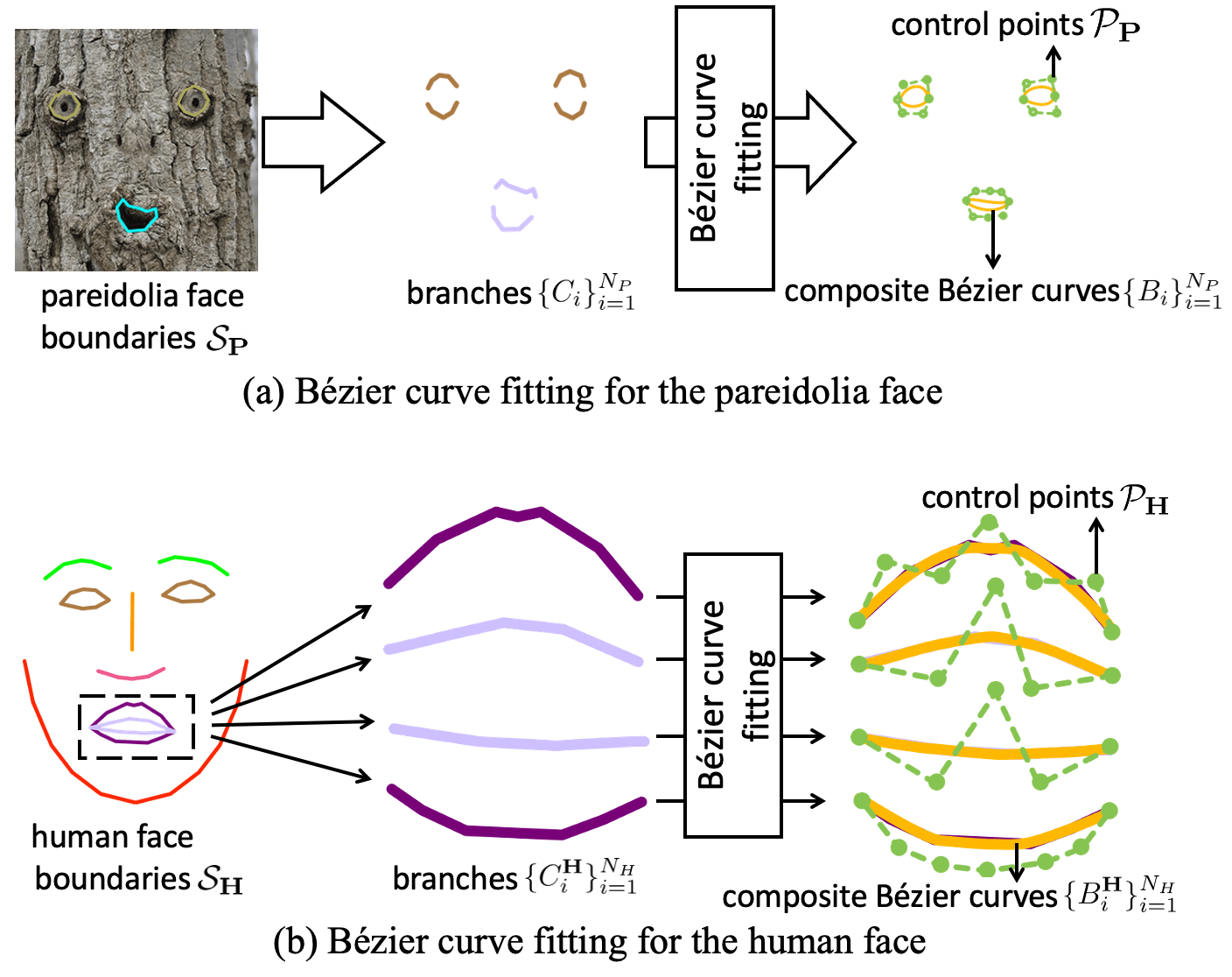}
    \caption{\textbf{Parametric Shape Modeling.} The boundaries of human face ($\mathcal{S}_{\mathbf{H}})$ and pareidolia face ($\mathcal{S}_{\mathbf{P}}$) are parameterized as control points of composite B\'{e}zier curves.
    For simplicity, we use the mouth as an example in $\mathcal{S}_{\mathbf{H}}$.
    }
    \label{fig:shape_motion}
    \vspace{-2mm}
\end{figure}

As shown in Fig.~\ref{fig:arch}, to represent the human face's shape independent of the face scale and rotation, we use the template alignment algorithm~\cite{segal2009generalized} to affine the human face $\mathbf{H}$ to the referred generic head model $\mathbf{F}$ of FLAME~\cite{FLAME:SiggraphAsia2017}. The aligned facial boundaries will be used for the following shape modeling.
In Fig.~\ref{fig:shape_motion}, we illustrate the procedure of shape modeling for human facial parts in detail.
First, by connecting inferred 68 3D landmarks~\cite{guo2020towards} of the human face $\mathbf{H}$ we obtain facial parts' boundaries $\mathcal{S}_{\mathbf{H}}=\{C_i^{\mathbf{H}}\}_{i=1}^{N}$ composed of $N_H$ branches.
For example, the mouth boundary can be divided into four branches: inner and outer contours of upper and lower lips.
Then, each branch $C_i^{\mathbf{H}}$ is fitted by a single $h_i$-order composite B\'{e}zier Curve $B_i^{\mathbf{H}}$ parameterized by $h_i+1$ control points $\mathcal{P}_{\mathbf{H}}^{B_i}=\{(\hat{x_j^i},\hat{y_j^i},\hat{z_j^i})\}_{j=0}^{h_i}$, where $(\hat{x_j^i},\hat{y_j^i},\hat{z_j^i})$ represents 3D coordinates of each control point.
Please refer to the \textit{supplementary material} about the single $n$-order composite B\'{e}zier curve fitting.
Thus, all branches of $\mathcal{S}_{\mathbf{H}}$ can be parameterized as control points $\mathcal{P}_{\mathbf{H}}=\bigcup_{i=1}^{N}\mathcal{P}^{B_i}_{\mathbf{H}}$.
At last, we conduct a similar procedure on the referred FLAME head model $\mathbf{F}$ and the boundaries $\mathcal{S}_{\mathbf{F}}=\{C_i^{\mathbf{F}}\}_{i=1}^{N}$ are parameterized as control points $\mathcal{P}_{\mathbf{F}}=\bigcup_{i=1}^{N}\mathcal{P}_{\mathbf{F}}^{B_i}=\bigcup_{i=1}^{N}\{(\bar{x_j^i},\bar{y_j^i},\bar{z_j^i})\}_{j=0}^{h_i}$.
For a pareidolia face, we manually label its boundaries $\mathcal{S}_{\mathbf{P}}=\{C_i^{\mathbf{P}}\}_{i=1}^{N_P}$ composed of $N_P$ branches and they are parameterized of control points $\mathcal{P}_{\mathbf{P}}$. 
By now, we get control points of facial parts' boundaries for $\mathbf{H}$, $\mathbf{F}$ and $\mathbf{P}$ as $\mathcal{P}_{\mathbf{H}}$, $\mathcal{P}_{\mathbf{F}}$, $\mathcal{P}_{\mathbf{P}}$ respectively, which is used for the following motion controller's calculation.

\noindent
\textbf{Motion Controller.}
Our motion representation extracted from human face, denoted as $\mathcal{M}_{\mathbf{S}}$, is defined as position of control points $\mathcal{P}_{\mathbf{H}}$ relative to $\mathcal{P}_{\mathbf{F}}$ as follows:

\vspace{-3mm}
\begin{equation}
    \mathcal{M}_{\mathbf{S}}=
    \bigcup_{i=1}^{N}\{(\frac{\hat{x_j^i}}{\bar{x_j^i}},\frac{\hat{y_j^i}}{\bar{y_j^i}},\frac{\hat{z_j^i}}{\bar{z_j^i}})\}_{j=0}^{h_i},
\end{equation}

\noindent
$\mathcal{M}_{\mathbf{S}}$ will animate boundaries of pareidolia faces $\mathcal{S}_{\mathbf{P}}$ and it is called as \textit{motion controllers} in Fig.~\ref{fig:shape_motion}.
In general, boundary branches of a pareidolia face are a subset of those of a human face, \textit{e.g.}, $N_P\le N$.
Because some facial parts of the pareidolia face such as nose and jawline are hard to define and only the observed facial parts are adopted, \textit{e.g.}, eyes and mouth in Fig.~\ref{fig:arch}.
First, for simplicity, we assume that the $i$-th boundary branch of $\mathbf{P}$ is corresponding to the $i$-th boundary branch of $\mathbf{H}$.
We parametrize the boundaries $\mathcal{S}_{\mathbf{P}}$ by control points $\mathcal{P}_{\mathbf{P}}=\bigcup_{i=1}^{N_P}\mathcal{P}_{B_i}=\bigcup_{i=1}^{N_P}\{(x_j^i,y_j^i,z_j^i)\}_{j=0}^{t_i}$, where $N_P$ is the branch number and $t_i$ is the curve order of the $i$-th branch.
Then, we note that the shape of curve $B_i$ (pareidolia face) might greatly differ from the shape of corresponding curve $B_i^{\mathbf{H}}$ (human face), \textit{e.g.}, the number of control points differs ($t_i\neq h_i$).
Thus, we uniformly remove (when $t_i<h_i$) or linearly interpolate (when $t_i>h_i$) the ordered motion controllers of the curve $B_i^{\mathbf{H}}$ in $\mathcal{M}_{\mathbf{S}}$ and denote the adapted motion controllers as $\mathcal{M}_{\mathbf{S}}^{e}$.
At last, facial parts of pareidolia face are animated by applying the motion controllers $\mathcal{M}_{\mathbf{S}}^{e}$ on their control points $\mathcal{P}_{\mathbf{P}}$ as follows:

\vspace{-3mm}
\begin{equation}
    \mathcal{P}_{\mathbf{P}}'=\mathcal{M}_{\mathbf{S}}^{e}\otimes\mathcal{P}_{\mathbf{P}}=\bigcup_{i=1}^{N_P}\{(\frac{\hat{x_j^i}}{\bar{x_j^i}}x_j^i,\frac{\hat{y_j^i}}{\bar{y_j^i}}y_j^i,\frac{\hat{z_j^i}}{\bar{z_j^i}}z_j^i)\}_{j=0}^{t_i},
    \label{eq:motion_control}
\end{equation}

\noindent
where $\otimes$ is point-wise dot product and $\mathcal{P}_{\mathbf{P}}'$ is the control points of the boundaries animated by motion $\mathcal{M}_{\mathbf{S}}^{e}$.

\subsection{Expansionary Motion Transfer}
\label{sec:motion_transfer_densify}
\begin{figure}[!t]
    \centering
    \includegraphics[width=0.95\linewidth]{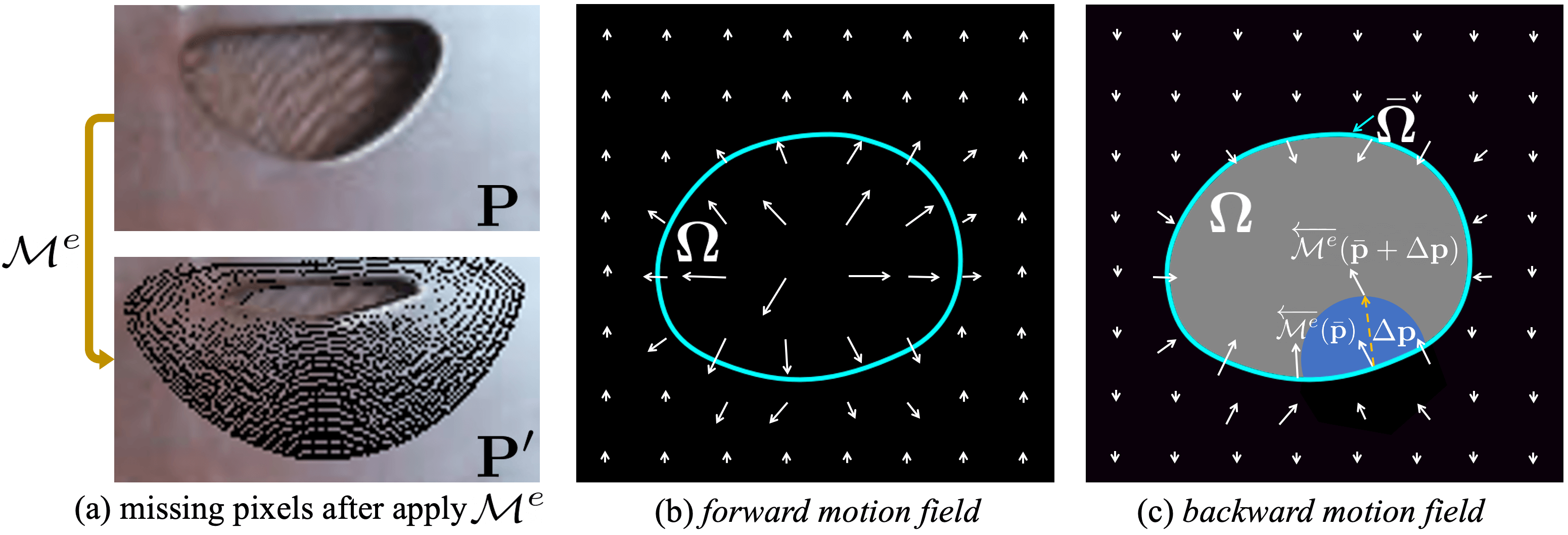}
    \caption{\textbf{First-order Motion Approximation.} (a) A practical example of missing pixels after directly applying the motion field. (b) \textit{Motion field} represents the motion of $G(\mathbf{P})\rightarrow G(\mathbf{P}')$. The pixel inside the hole $\mathbf{\Omega}$ move outward. (c) \textit{Inverse motion field} misses valid values inside $\mathbf{\Omega}$ and the proposed First-order Motion Approximation is used to approximate them.
    }
    \label{fig:motion_approx}
    \vspace{-2mm}
\end{figure}

Now we transfer the motion at the facial boundaries by animated control points of composite B\'{e}zier curves in the pareidolia face.
However, the transferred motion is local and a global motion of the whole face is required to animate a pareidolia face.
Thus, we develop a Motion Spread strategy to expand the motion at the boundaries to the whole face.
Moreover, we find that the texture animated by the expanded motion contains missing pixels as shown in Fig.~\ref{fig:motion_approx} (a).
We will detail the cause and then propose the First-order Motion Approximation to address it.

\noindent
\textbf{Motion Spread.}
The motion at curve $B_i$, denoted as $\mathcal{M}_{B_i}^{e}$, is defined by the optical flow map of points on composite B\'{e}zier Curves parameterized by $\mathcal{P}_{\mathbf{P}}$ and $\mathcal{P}_{\mathbf{P}}'$.
We call the facial motion at each boundary branch of facial parts, \textit{e.g.} $\mathcal{M}_{B_i}^{e}$, as the \textit{motion seed}.
Then $\{\mathcal{M}_{B_i}^{e}\}_{i=1}^{N_P}$ is called as \textit{motion seeds} as shown in Fig.~\ref{fig:arch}.
Each composite B\'{e}zier curve $B_i$ is related to a motion seed $\mathcal{M}_{B_i}^{e}$ in the pareidolia face $\mathbf{P}$.
Note that the motion seeds only define a very local motion at the facial boundaries, we develop a \textit{Motion Spread} strategy to derive the facial motion of the whole pareidolia face.

\begin{figure}[ht]
    \centering
    \includegraphics[width=0.95\linewidth]{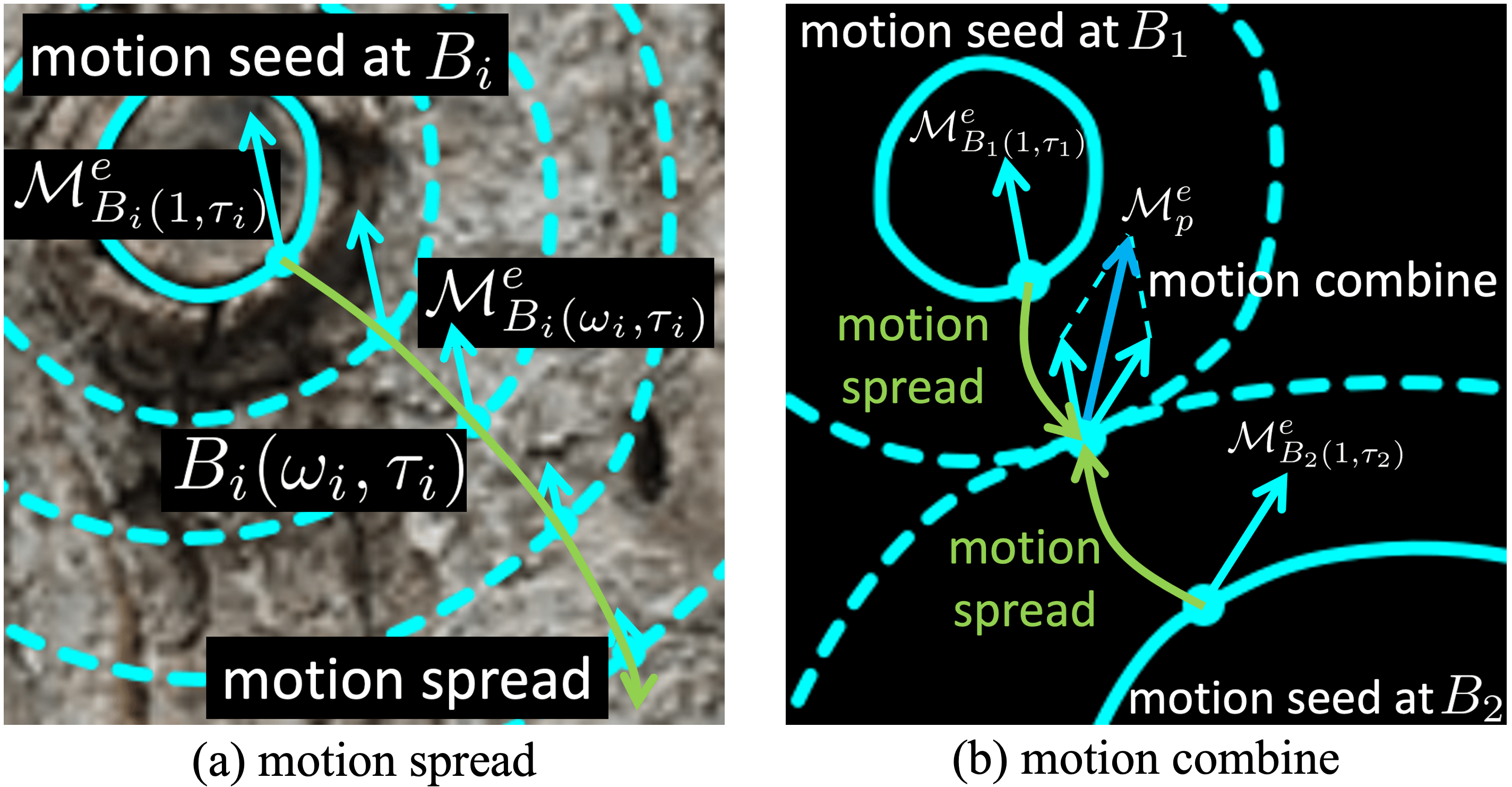}
    \caption{\textbf{(a) motion spread.} A motion $\mathcal{M}_{B_i(1,\tau_i)}^{e}$ of the motion seed at $B_i$ spreads to $B_i(\omega_i,\tau_i)$ as motion $\mathcal{M}_{B_i(\omega_i,\tau_i)}$. \textbf{(b) motion combine.} At pixel $\mathbf{p}$, we combine spreaded motion from motion seeds at $B_1,B_2$.
    }
    \label{fig:motion_spread_combine}
    \vspace{-2mm}
\end{figure}

Our Motion Spread strategy decays motion seeds along the directions orthogonal with their composite B\'{e}zier curves as shown in Fig.~\ref{fig:motion_spread_combine} (a). Each motion seed $\mathcal{M}_{B_i}^{e}$ represents the motion of all points on curve $B_i$.
We expand the curve $B_i$ to different scales to cover its neighboring area. 
If a pixel position $\mathbf{p}$ locates at the relative position $\tau_i\ (\tau_i\in[0,1])$ of $\omega_i$-time scaled curve $B_i$, we can represent it as $\mathbf{p}=B_i(\omega_i,\tau_i)$.
Then, the decayed motion $\mathcal{M}_{B_i(\omega_i,\tau_i)}^{e}$ from the motion seed $\mathcal{M}_{B_i}^{e}$ can be written as follows:

\vspace{-3mm}
\begin{equation}
    \mathcal{M}_{B_i(\omega_i,\tau_i)}^{e} = \lambda(\omega_i)\cdot \mathcal{M}_{B_i(1,\tau_i)}^{e},\ 
    \mathcal{M}_{B_i(1,\tau_i)}^{e}\in \mathcal{M}_{B_i}^{e},
\end{equation}

\noindent
where $\cdot$ means scalar multiplication and the motion decay factor $\lambda(\omega_i)$ is determined by $\omega_i$ as presented in the \textit{supplementary material}.
\noindent
One pixel $\mathbf{p}$ in the pareidolia face might receive decayed motion from several motion seeds.
Therefore, we calculate the motion at $\mathbf{p}$ by motion combine.
The motion at $\mathbf{p}$, denoted as $\mathcal{M}_{\mathbf{p}}^{e}$, is the combination of these decay motion as shown in Fig.~\ref{fig:motion_spread_combine} (b).

To animate the pareidolia face $\mathbf{P}$, a global motion for the whole pareidolia face is built from the motion seeds through our proposed Motion Spread strategy.
Such a global motion is constituted by the motion of $\mathcal{M}_{\mathbf{p}}^{e}$ for all $\mathbf{p}\in G(\mathbf{P})$, where $G(\cdot)$ is a function that returns the pixel grid of the input image.
We call it as \textit{motion field} and denote it as $\mathcal{M}^{e}=\{\mathcal{M}_{\mathbf{p}}^{e}\}_{\mathbf{p}\in G(\mathbf{P})}$.
Then, for a pixel at $\mathbf{p}$ in the pareidolia face, the motion field can animate it to a new location $\mathcal{M}_{\mathbf{p}}^{e} + \mathbf{p}\ (\forall \mathbf{p}\in G(\mathbf{P}),\ \mathcal{M}_{\mathbf{p}}^{e}\in\mathcal{M}^{e})$ by the motion field $\mathcal{M}^{e}$.

\noindent
\textbf{First-order Motion Approximation.}
\label{sec:motion_approx}
The motion field $\mathcal{M}^{e}$ can be used to animate the pareidolia face $\mathbf{P}$ as the reenacted face $\mathbf{P}'$.
If we regard $\mathcal{M}^{e}$ as a function $G(\mathbf{P})\rightarrow G(\mathbf{P}')$, then it is neither an injection nor a surjection since multiple pixel locations in $G(\mathbf{P})$ might be mapped to one pixel location in $G(\mathbf{P}')$.
Thus, directly using $\mathcal{M}^{e}$ to animate $\mathbf{P}$ will cause some missing pixels in $\mathbf{P}'$ as shown in Fig.~\ref{fig:motion_approx} (a).
To solve this problem, we introduce the inverse function of $\mathcal{M}^{e}$, denoted as $\overleftarrow{\mathcal{M}^{e}}:G(\mathbf{P}')\rightarrow G(\mathbf{P})$ since $\overleftarrow{\mathcal{M}^{e}}$ do not have valid function value at the locations of missing pixels.
Thus, our goal is to inpaint the $\overleftarrow{\mathcal{M}^{e}}$ and we propose our \textit{First-order motion approximation}.

To illustrate the First-order Motion Approximation, we take a typical case in Fig.~\ref{fig:motion_approx} (b) where all pixels in the area $\mathbf{\Omega}$ move outward.
At first, we call $\overleftarrow{\mathcal{M}^{e}}$ as \textit{inverse motion field} since it defines the pixel movements opposite to that of the motion field.
Then, in Fig.~\ref{fig:motion_approx} (c), $\overleftarrow{\mathcal{M}^{e}}$ does not contain valid function value in the area $\mathbf{\Omega}\subset G(\mathbf{P}')$.
At last, since close pixels will have similar motion, we propose the First-order Motion Approximation for $\overleftarrow{\mathcal{M}^{e}}$ to spread facial motion from the area boundary $\bar{\mathbf{\Omega}}$ to the area $\mathbf{\Omega}$.
Specially, We compute the first-order Taylor expansion of $\overleftarrow{\mathcal{M}^{e}}$ around the area boundary pixel $\bar{\mathbf{p}}\in\bar{\mathbf{\Omega}}$ as follows:

\footnotesize
\begin{equation}
    \overleftarrow{\mathcal{M}^{e}}(\bar{\mathbf{p}}+\Delta\mathbf{p})=\overleftarrow{\mathcal{M}^{e}}(\bar{\mathbf{p}})+(\frac{d}{d\mathbf{p}}\overleftarrow{\mathcal{M}^{e}}(\mathbf{p})|_{\mathbf{p}=\bar{\mathbf{p}}})\Delta\mathbf{p}+o(||\Delta\mathbf{p}||),
    \label{eq:first_order}
\end{equation}
\normalsize

\noindent
where $\bar{\mathbf{p}}+\Delta\mathbf{p}$ is a nearby pixel of the boundary $\bar{\mathbf{p}}$.
We omit $o(||\Delta\mathbf{p}||)$ of Eq.~(\ref{eq:first_order}) to approximate $\overleftarrow{\mathcal{M}^{e}}(\bar{\mathbf{p}}+\Delta\mathbf{p})$.
We conduct Eq.~(\ref{eq:first_order}) for all missing pixels in $\overleftarrow{\mathcal{M}^{e}}$ and the enhanced \textit{inverse motion field} is denoted as $\overleftarrow{\mathcal{M}^{E}}$.
The motion approximation along $\Delta\mathbf{p}$ can be viewed in Fig.~\ref{fig:motion_approx} (c).

\subsection{Unsupervised Texture Animator}
\label{sec:texture_syn}
While we get the inverse motion field, the final step is to synthesize the image result conditioned on the raw pareidolia face $\mathbf{P}$ and the inverse motion field.
Reviewing the absence of large-scale datasets and annotations for pareidolia face, we propose an AutoEncoder based Unsupervised Texture Animator.
Specifically, we first train a simple AutoEncoder with only natural images without any annotation.
The trained AutoEncoder can be seen as a texture reconstructor by now, which can reconstruct the input texture but cannot animate it.
To this end, we design a Feature Deforming Layer, which is coupled with the AutoEncoder network, to transfer the motion to texture progressively.
Note that Feature Deforming Layer is only used in the inference stage, which makes the training unsupervised and enjoys the diversity of large-scale datasets of natural images.

\begin{figure}[!t]
    \centering
    \vspace{-6mm}
    \includegraphics[width=0.95\linewidth]{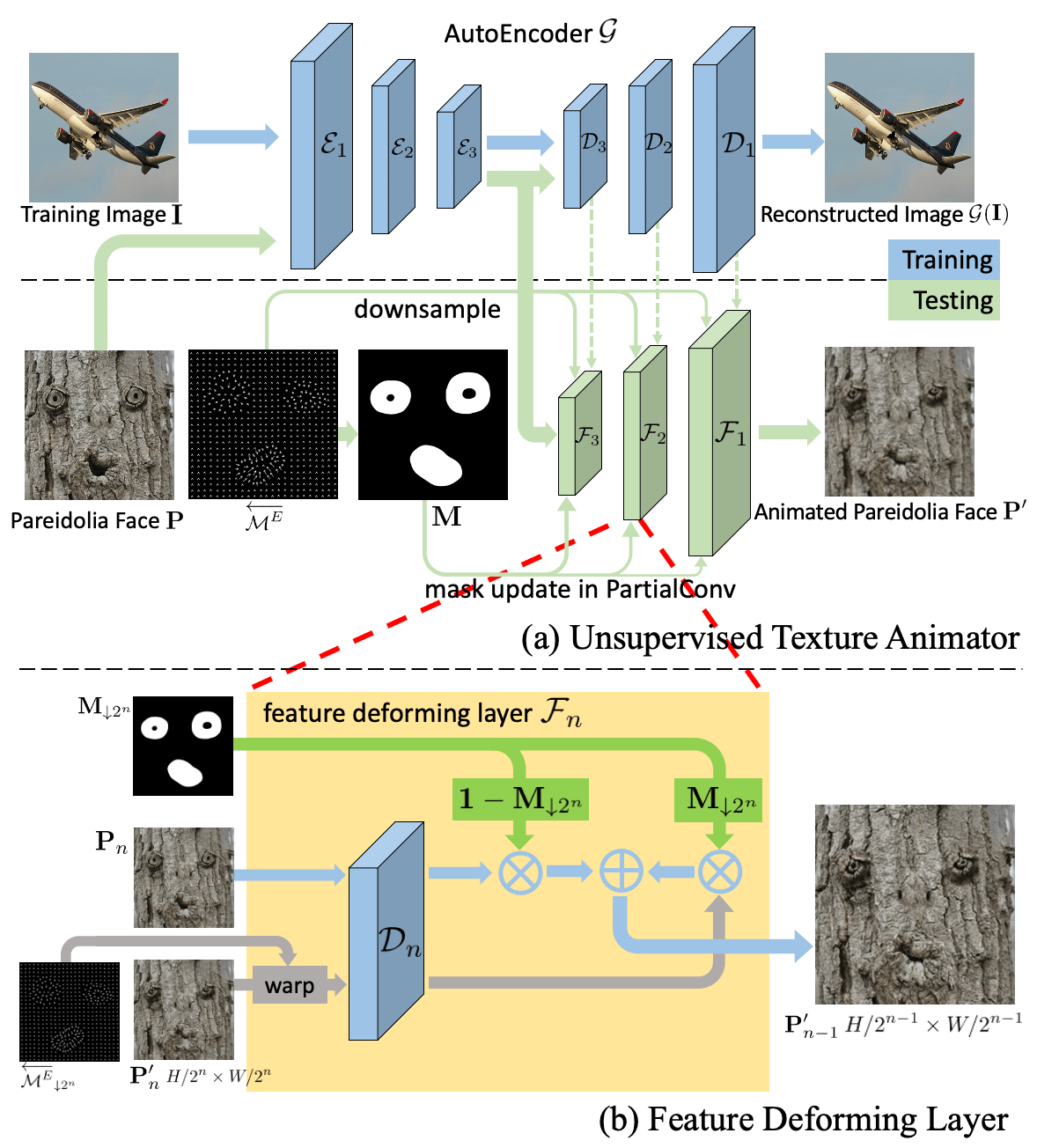}
    \caption{\textbf{(a) Unsupervised Texture Animator.} Training and testing phases of our Unsupervised Texture Animator.
    \textbf{(b) Feature Deforming Layer.}
    For simplifying the understanding, we visualize the feature maps $\mathbf{P}_n,\mathbf{P}_n',\mathbf{P}_{n-1}'$ using the tree texture images.
    }
    \label{fig:texture_animator}
    \vspace{-2mm}
\end{figure}

\noindent
\textbf{Unsupervised AutoEncoder.}
\label{sec:texture_animator}
We train an unsupervised AutoEncoder $\mathcal{G}$ to extract image features at different scales as shown in Fig.~\ref{fig:texture_animator}.
During the training phase, an image $\mathbf{I}$ is fed into $\mathcal{G}$ to produce reconstructed image $\mathcal{G}(\mathbf{I})$. 
We apply $l_1$ reconstruction loss $L_{rec}$ and perceptual loss~\cite{Johnson2016PL} $L_{vgg}$ on $\mathbf{I}$ and $\mathcal{G}(\mathbf{I})$.
The loss function of $\mathcal{G}$ is $L_{\mathcal{G}}=\alpha_1 L_{rec}+\alpha_2 L_{vgg}$, where $\alpha_1$ and $\alpha_2$ are set empirically.
We put the network details in the \textit{supplementary material}.

\noindent
\textbf{Feature Deforming Layer.}
The pareidolia image features $\mathbf{P}_n$ of the scale $H/{2^n}\times W/{2^n}$ can be retrieved from layer $\mathcal{D}_n$ in the pretrained $\mathcal{G}$.
During the testing phase, in Fig.~\ref{fig:texture_animator}, we design the \textit{Feature Deforming Layer} $\mathcal{F}_n$ to warp the synthesized features $\mathbf{P}_n'$ by the downsampled motion field $\overleftarrow{\mathcal{M}^{E}}_{\downarrow 2^n}$ and refine it by $\mathcal{D}_n$.
Thus, the texture is progressively synthesized by $\mathcal{F}_{3},\mathcal{F}_{2},\mathcal{F}_{1}$.
At last, since some pixels of $\mathbf{P}$ do not move in $\mathbf{P}'$.
A 0-1 motion mask $\mathbf{M}$ is calculated based on $\overleftarrow{\mathcal{M}^{E}}$ to keep the texture at these pixels unchanged.
The motion mask $\mathbf{M}$ is defined as $\mathbf{M}=\mathbbm{1}(\overleftarrow{\mathcal{M}^{E}}-\mathcal{M}^{I})$, 
where $\mathbbm{1}$ is an indicator function that returns 0 if the input is 0 and returns 1 elsewise, and $\mathcal{M}^{I}$ is an identity motion field that do not move any pixel of $\mathbf{P}$. Thus, the texture refinement of layer $\mathcal{F}_{n}$ is written as:

\vspace{-3mm}
\begin{equation}
    \mathbf{P}_{n-1}'=\mathcal{F}_{n}(\mathbf{P}_{n},\mathbf{P}_{n}',\overleftarrow{\mathcal{M}^{E}}_{\downarrow 2^n},\mathbf{M}_{\downarrow 2^n}),
    \label{eq:texture_refine}
\end{equation}

\begin{figure*}[ht!]
    \centering
    \vspace{-5mm}
    \includegraphics[width=0.9\linewidth]{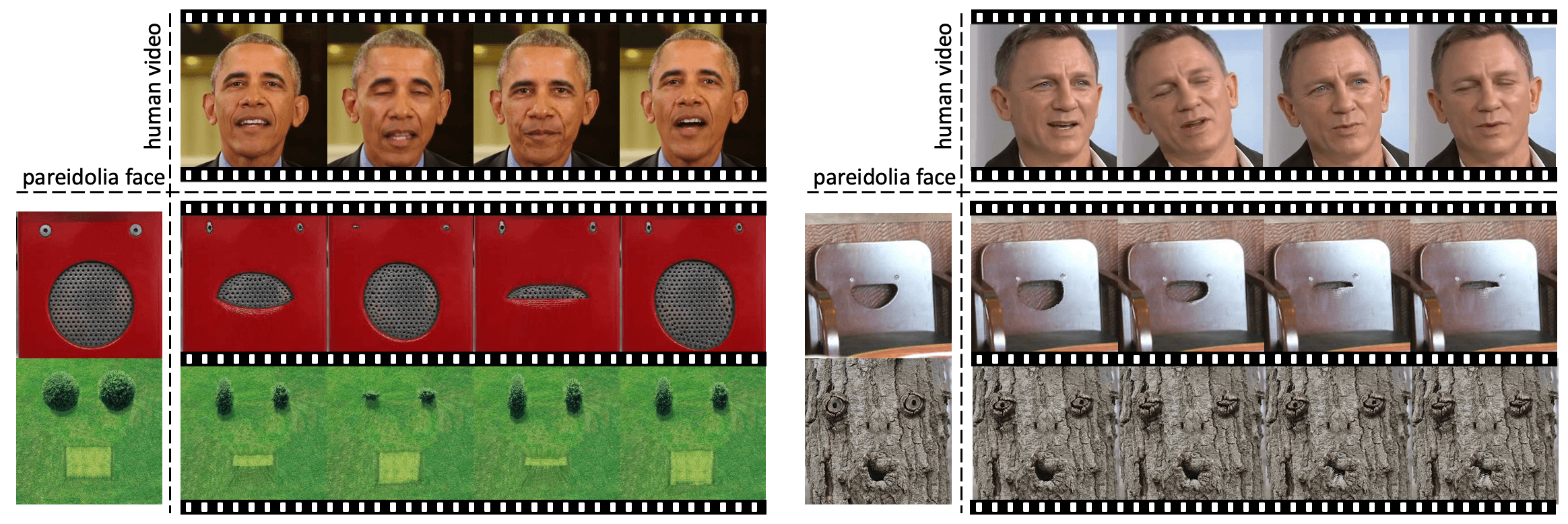}
    \caption{\textbf{Pareidolia Face Reenactment.} In each block, we use the human face portrait video in the first row to drive pareidolia faces in the second and third row. Our method mainly focuses on facial motion transferring at the mouth and eyes.
    }
    \label{fig:pare_face_reenact}
    \vspace{-2mm}
\end{figure*}

\noindent
where $\mathbf{M}_{\downarrow 2^n}$ is obtained by downsampling the mask $\mathbf{M}$ ($2^n$ is the scale factor) through the mask update method in PartialConv~\cite{liu2018image}. In Eq.~(\ref{eq:texture_refine}), $\mathbf{P}_{3}'$ is our coarsest texture produced by the encoder of $\mathcal{G}$.
By the progressive warp and refine, we get $\mathbf{P}_{0}'=\mathbf{P}'$ as our final synthesized pareidolia face with the same motion of the human face video.

\section{Experiments}
We show qualitative and quantitative results on the generated videos of pareidolia faces to demonstrate the performance of our reenactment method for pareidolia faces and the effectiveness of the proposed components.

\noindent
\textbf{Datasets.} During the training phase, the AutoEncoder $\mathcal{G}$ is trained on the COCO2017 dataset~\cite{lin2014microsoft}.
During the testing phase, the human portrait videos include videos from \textit{Obama Weekly Address}~\cite{suwajanakorn2017synthesizing} and CelebVox2~\cite{chung2018voxceleb2}. Moreover, we collect a dataset PareFace, which includes $1,000$ pareidolia faces to facilitate future researches on this topic. More samples can be viewed in our \textit{supplementary material}.

\noindent
\textbf{Metric.} We evaluate videos of the animated pareidolia faces in terms of textures and the shape/motion of facial parts.
We use IS~\cite{barratt2018note}, FID~\cite{heusel2017gans} to evaluate the synthesized texture quality.
Due to the lack of metrics about evaluating the shape and motion differences between human and pareidolia faces. We design the following metrics to evaluate the shape similar and motion accuracy:
1) \textbf{Shape similarity (S-Sim).} Following ~\cite{ramachandran2015combined}, we use the eccentricity histogram of a shape as its descriptor. The cosine distance is used to measure the similarity between two shapes.
2) \textbf{Close-open accuracy (CO-Acc).} To measure the extreme motion accuracy of the mouth and eyes, we compare their open/close status. The CO-Acc is defined as the average difference of the mouth/eyes open ratio between input human and animated pareidolia faces, where the open ratio is expressed as a percentage of the maximum height of the mouth/eyes.
3) \textbf{Motion accuracy (M-Acc).} To measure the overall motion accuracy of the mouth and eyes, we compare their tendencies of becoming larger/smaller. We use 0-1 flags to denote if the area becomes larger or smaller in the next frame. The flag serves as a motion indicator and we compare its average differences between human and pareidolia faces.

\subsection{Pareidolia Face Reenactment}
\begin{figure}[!t]
    \centering
    \includegraphics[width=0.95\linewidth]{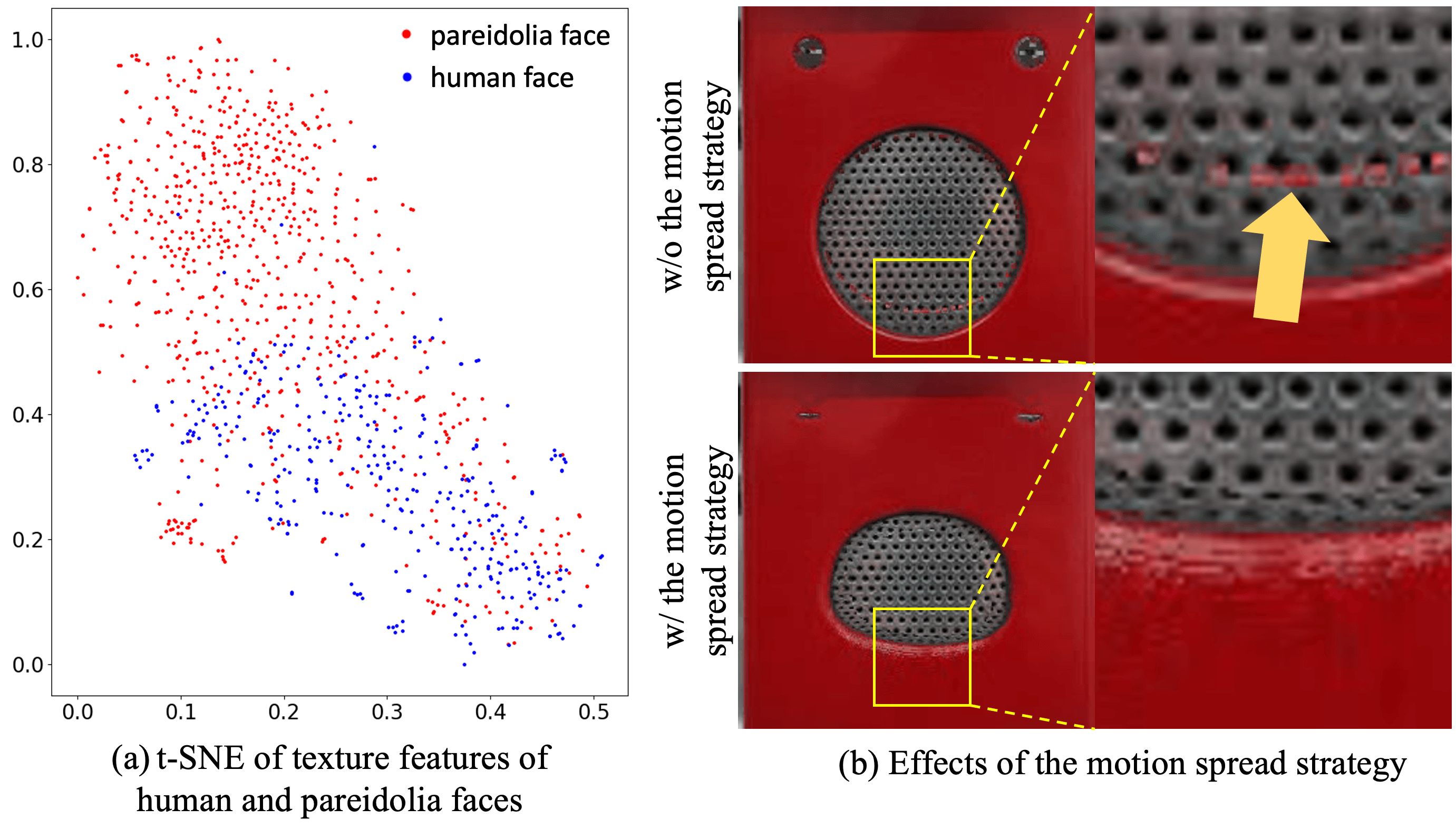}
    \caption{\textbf{(a) t-SNE of texture features of human and pareidolia faces.} 
    The texture feature is extracted by VGG16~\cite{Simonyan2015VeryDC} pre-trained on ImageNet.
    \textbf{(b) Effect of Motion Spread Strategy.}
    Comparison of animated results before and after applying the Motion Spread strategy.
    }
    \label{fig:tsne_spread}
    \vspace{-2mm}
\end{figure}

We use a human portrait video to drive a given pareidolia face and present visual results in Fig.~\ref{fig:pare_face_reenact}.
It can be seen that the generated pareidolia face imitates the motion of the input human face at the mouth and eyes areas, even the subtle size changing.
Benefits from our parametric shape modeling, the prominent motion at facial parts is transferred from the human to pareidolia face.
Our Motion Spread strategy makes it possible to animate the area around facial parts and make the whole pareidolia face looks more lively.
Even large texture discrepancy exists between human and pareidolia faces as shown in Fig.~\ref{fig:tsne_spread} (a), where the distribution of pareidolia faces' texture is more discrete than that of human faces.
We recommend to view reenactment results in the \textit{supplementary video}.

\subsection{Ablation Study}
In this section, we present an ablation to evaluate the effectiveness of our proposed components.
First, we manually label ``landmarks'' for pareidolia faces and use them to replace the control points of composite B\'{e}zier curves to validate the effect of our parametric shape modeling (Sec.~\ref{sec:abl_bezier_ldmk}).
Then, we show the decisive role that the Motion Spread strategy plays in the pareidolia face animation (Sec.~\ref{sec:motion_spread}).
In addition, we compare the texture quality improvement brought by our First-order Motion Approximation (Sec.~\ref{sec:approx}).
At last, we progressively add our proposed Feature Deforming Layers ($\mathcal{F}_1,\mathcal{F}_2,\mathcal{F}_3$) in the Unsupervised Texture Animator to visualize the progressively refined textures (Sec.~\ref{sec:abl_texture_animate}).
\vspace{-2mm}
\subsubsection{Composite B\'{e}zier Curve v.s. Landmarks}
\label{sec:abl_bezier_ldmk}
\begin{figure}[!t]
    \centering
    \includegraphics[width=0.95\linewidth]{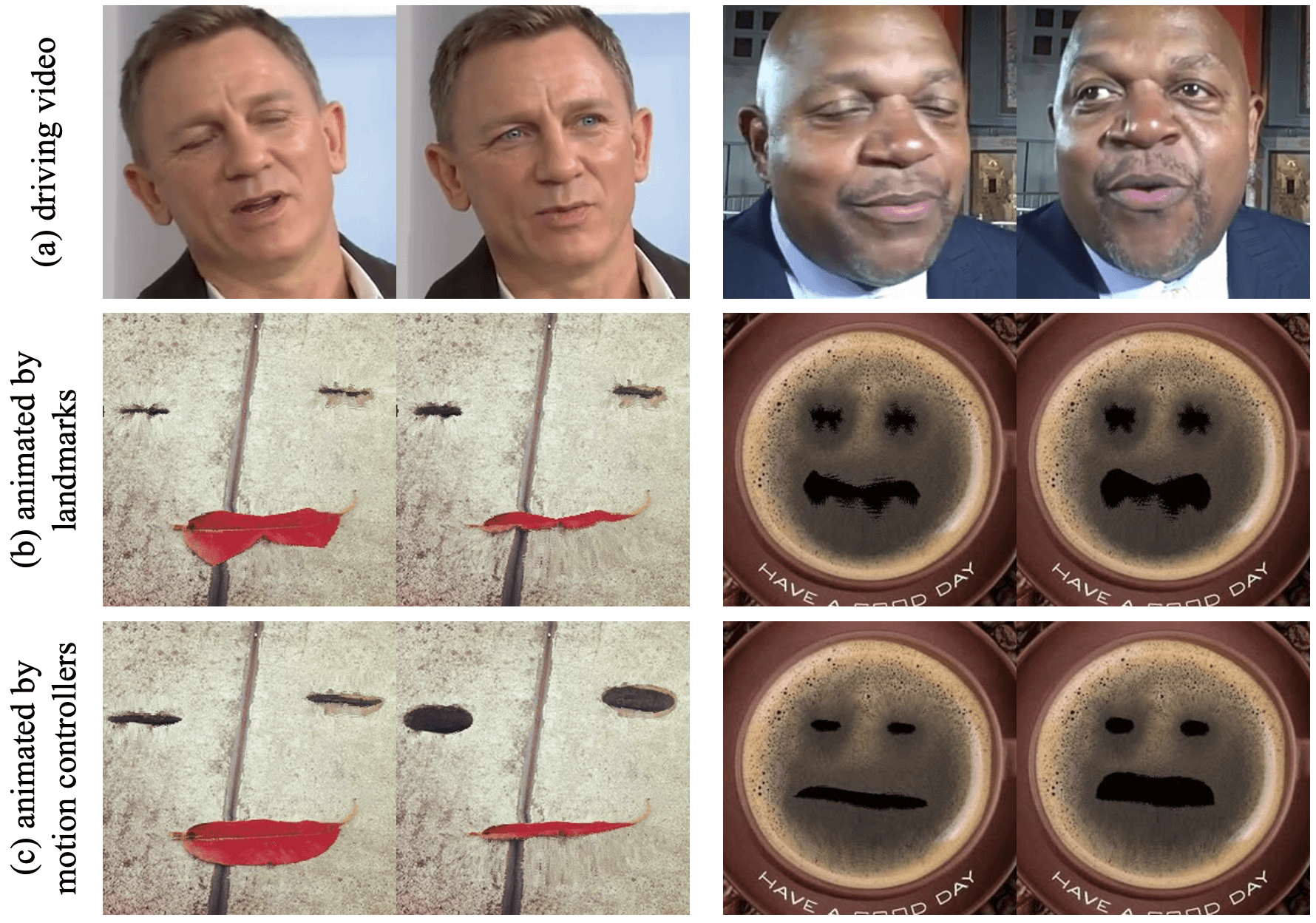}
    \caption{\textbf{Composite B\'{e}zier Curves v.s. Landmarks.}
    Our proposed representation \textit{motion controllers} inspired by composite B\'{e}zier curve is better than landmarks in reenacting pareidolia faces.
    }
    \label{fig:bcurve_ldmk}
    \vspace{-2mm}
\end{figure}

To valid the superiority of incorporating composite B\'{e}zier curves,
we compare our method with the one that replaces the control points of composite B\'{e}zier curve with the manually labeled ``landmarks'' for pareidolia faces. We present the qualitative results in Fig.~\ref{fig:bcurve_ldmk}.
We can see that the facial parts' global shape is broken when landmarks are applied while our method preserves them well.
Also, the reenactment results produced by our method also imitate the facial motion of the human face better (the eyes' motion of the left man in Fig.~\ref{fig:bcurve_ldmk}).
To quantitatively compare the shape similar, motion accuracy and image quality, we present Tab.~\ref{tab:shape_motion_visual} (a) and Tab.~\ref{tab:shape_motion_visual} (b) that compare the results of driving by landmarks and composite B\'{e}zier curves.
Compare with landmarks, modeling facial parts by composite B\'{e}zier curves are better at preserving the facial parts' global shapes, imitating the motion of human faces and the synthesized image visual quality.

\begin{table}[ht]
    \small
    \centering
    \caption{\textbf{(a) Shape similarity and motion accuracy of Motion Controllers v.s. Landmarks.}
    In the table, 'm' means mouth and 'e' means eyes.
    \textbf{(b) Visual Quality Comparison.}
    The IS and FID of images synthesized by different settings are compared.
    In the `landmark', we use labeled landmarks of pareidolia faces instead of composite B\'{e}zier curves.
    In the `w/o motion appr.', we do not apply the First-order Motion Approximation.
    In the `Ours', the Unsupervised Texture Animator applies the network structure $\mathcal{F}_2,\mathcal{F}_2,\mathcal{F}_3$.
    }
    \resizebox{1.0\linewidth}{!}
    {
    \begin{tabular}{c|c|c}
        \Xhline{1pt}
        \textbf{(a)} Method & landmark & \makecell{composite\\B\'{e}zier curve}\\
        \Xhline{1pt}
        S-Sim(m) & 0.43 & \textbf{0.75} \\
        S-Sim(e) & 0.55& \textbf{0.82} \\
        \Xhline{1pt}
        CO-Acc(m) & 0.52 & \textbf{0.76} \\
        CO-Acc(e) & 0.71 & \textbf{0.82} \\
        \Xhline{1pt}
        M-Acc(m) & 0.77 & \textbf{0.84} \\
        M-Acc(e) & 0.80 & \textbf{0.89} \\
        \Xhline{1pt}
    \end{tabular}
    \quad
    \begin{tabular}{c|c|c}
        \Xhline{1pt}
        \textbf{(b)} Method & IS & FID \\
        \Xhline{1pt}
        landmark & 8.21 & 13.1 \\
        $\mathcal{F}_1,\mathcal{D}_2,\mathcal{D}_3$ & 8.79 & 12.9 \\
        $\mathcal{F}_2,\mathcal{F}_2,\mathcal{D}_3$ & 8.89 & 12.5 \\
        \makecell{w/o motion appr.} & 9.17 & 12.3 \\
        \Xhline{1pt} 
        \textbf{Ours} & \textbf{9.22} & \textbf{12.3} \\
        \Xhline{1pt}
    \end{tabular}
    \label{tab:shape_motion_visual}
    }
    \vspace{-2mm}
\end{table}

\vspace{-2mm}
\subsubsection{Effect of the Motion Spread Strategy}
\label{sec:motion_spread}
We propose the Motion Spread strategy to obtain the motion filed that defines the global motion of the pareidolia face from the motion seeds that only define the motion of facial parts' boundaries.
As shown in Fig.~\ref{fig:tsne_spread} (b), only the pixels at facial boundaries are animated without the Motion Spread strategy, which presents failed animation results.
Benefiting from the Motion Spread strategy, we can animate the whole pareidolia face.

\vspace{-2mm}
\subsubsection{Effect of the Motion Approximation}
\label{sec:approx}
\begin{figure}[!t]
    \centering
    \includegraphics[width=0.95\linewidth]{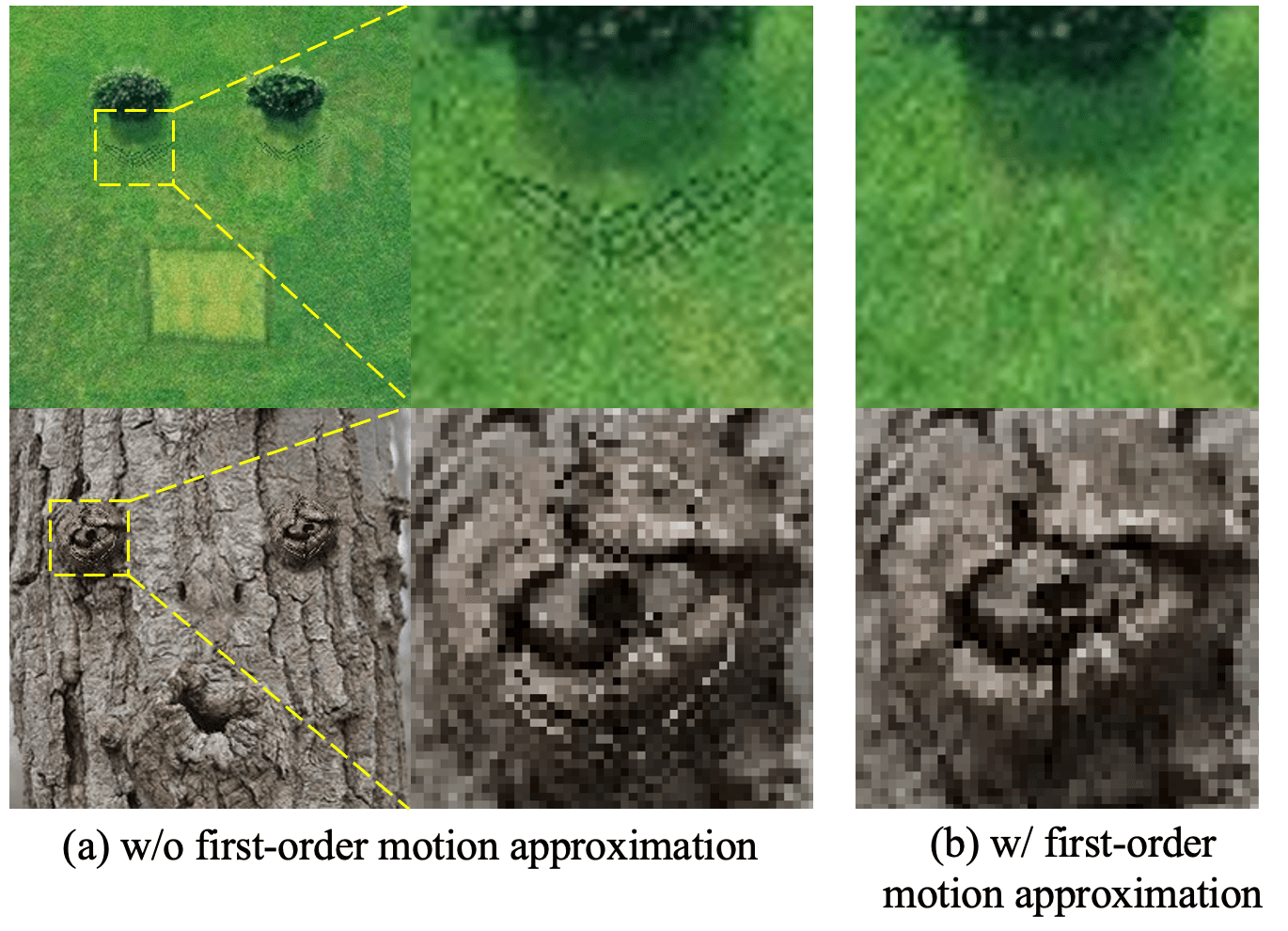}
    \vspace{-2mm}
    \caption{\textbf{Effect of First-order Motion Approximation.} 
    The texture synthesized by our Unsupervised Texture Animator when our proposed First-order Motion Approximation is applied (b) or not (a).
    }
    \label{fig:first_order_approx}
    \vspace{-2mm}
\end{figure}

Our First-order Motion Approximation is designed to approximate the missing motion in the inverse motion field.
Thus, the quality of the synthesized textures is improved after introducing this motion approximation method, as shown in Fig.~\ref{fig:first_order_approx}.
Some pixel-level visual artifacts in Fig.~\ref{fig:first_order_approx} (a) are caused by the missing value in the inverse motion field, thus the original pixels are remained.
They are obviously eliminated after applying our First-order Motion Approximation.
We also present the quantitative results of image quality in Tab.~\ref{tab:shape_motion_visual} (b).
Both IS and FID are slightly improved after applying our motion approximation method.

\vspace{-2mm}
\subsubsection{Effect of the Feature Deforming Layer}
\label{sec:abl_texture_animate}
\begin{figure}[!t]
    \centering
    \includegraphics[width=0.95\linewidth]{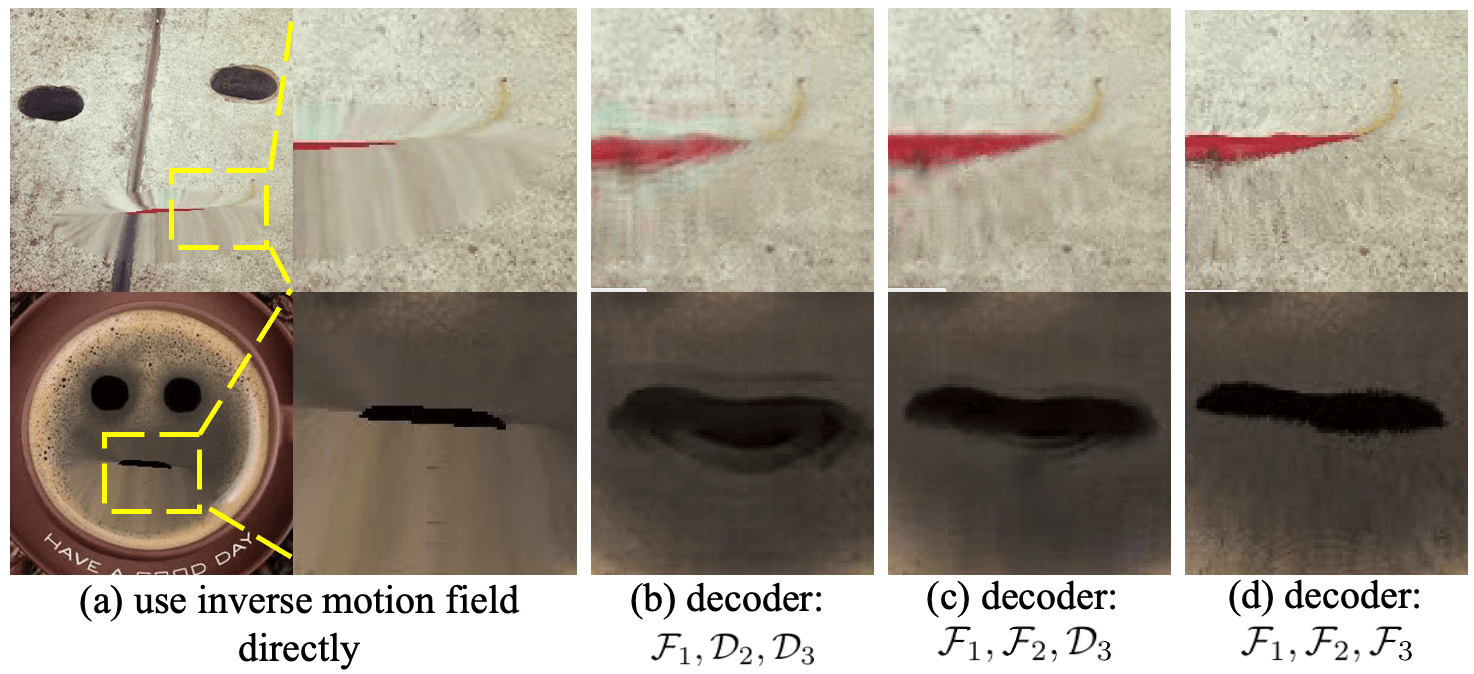}
    \caption{\textbf{Effect of the Feature Deforming Layers.} (a) We directly use the inverse motion field to animate the pareidolia face.
    (b-d) The texture quality of animated pareidolia faces is progressively improved by our Feature Deforming Layers.
    }
    \label{fig:layer_wise_refine}
    \vspace{-2mm}
\end{figure}

Directly applying the inverse motion field usually induce blurred synthesized textures.
Our Feature Deforming Layers in the Unsupervised Texture Animator benefits the texture synthesis by progressively warp and refine the input pareidolia faces.
We present the textures synthesized by each Feature Deforming Layer in Fig.~\ref{fig:layer_wise_refine} where we can see that the synthesized textures become more and more clear after we add $\mathcal{F}_1,\mathcal{F}_2,\mathcal{F}_3$ layer by layer.
We also present the quantitative results of image quality in terms of IS and FID in Tab.~\ref{tab:shape_motion_visual} (b) where both metrics are improved if more Feature Deforming Layers are applied.

\vspace{-3mm}

\section{Discussion}
In this paper, we make the first attempt on pareidolia face reenactment, which might benefit the cartoon production and mixed reality in the future. We present a Parametric Unsupervised Reenactment Algorithm to tackle this challenging problem and demonstrate superior reenactment results in the experiments.

Pareidolia face reenactment is an extremely challenging problem that might fail when meet extreme head poses, complex shapes and textures of the pareidolia faces.
Our method mainly focuses on transferring the motion of eyes and mouth from human faces to the frontal pareidolia faces.
We leave the motion transferring of other facial parts (\eg nose and facial muscles), the head movement and animating non-frontal pareidolia faces as our future works.  

\section*{Acknowledgments}
This research was conducted in collaboration with SenseTime. This work is partially funded by Beijing Natural Science Foundation (Grant No. JQ18017), Youth Innovation Promotion Association CAS (Grant No. Y201929), and National Natural Science Foundation of China (Grant No. U20A20223). This work is supported by A*STAR through the Industry Alignment Fund - Industry Collaboration Projects Grant.

{\small
\bibliographystyle{ieee_fullname}
\bibliography{egbib}
}

\newpage

\section*{Appendix}

\appendix

\section{More Details of the Methodology}
\label{sec:details}

\subsection{$n$-order Composite Bézier Curve Fitting}
For a frame $\mathbf{H}$ from human portrait video, we assume that there are $N_H$ branches in its boundaries $\mathcal{S}_{\mathbf{H}}=\{C_i\}_{i=1}^{N_H}$.
We use $n_i$-order composite Bézier curve to fit the boundary branch $C_i$ and denote the estimated composite Bézier curve as $B_i$.
The overall optimization problem can be written as follows:

\begin{equation}
    {\min} \sum_{i=1}^{N_H}||C_i-B_i||^2,
    \label{eq:original_problem}
\end{equation}

\noindent
where a composite Bézier curve $B_i$ is composited by $N_i$ vanilla  Bézier curves and we denote these vanilla Bézier curves as $B_{ij}\ (i\le N_S,j\le N_i\})$.
The composite Bézier curve $B_i$ are splitted to $N_i$ Bézier curves by $N_i-1$ \textit{joints}.
There are also $N_i-1$ joints on the boundary branch $C_i$ that correspond to the joints on $B_i$. Thus, $C_i$ is splitted as $C_{ij}\ (i\le N_S,j\le N_i\})$.
Thus, the overall optimization problem is formed as:

\begin{equation}
    \begin{array}{c}
    {\min} \sum_{i=1}^{N_H}||C_i-B_i||^2=
    {\min} \sum_{i=1}^{N_H}\sum_{j=1}^{N_i}||C_{ij}-B_{ij}||^2
    \\
    \Leftrightarrow
    {\min}||C_{ij}-B_{ij}||^2,\ \forall i,j,
    \end{array}
    \label{eq:optimize}
\end{equation}

\noindent
where $B_{ij}$ is a vanilla Bézier curve and in Eq.~(\ref{eq:optimize}) the original optimization problem is splitted into $N_H\times N_i$ independent optimization problems.
Thus, we consider the new optimization problems ${\min}||C_{ij}-B_{ij}||^2,\ \forall i,j$.
We omit the subscripts $i,j$ for simplicity.

According to the definition of $n$-order Bézier curve, a Bézier curve $B$ can be rewritten as follows,

\begin{equation}
    B(\tau)=\sum_{k=0}^{n}{n\choose k}\tau^{k}(1-\tau)^{n-k}P_k,
    \label{eq:bezier_curve}
\end{equation}

\noindent
where $\tau\in[0,1]$ represents the relative position of point $B(\tau)$ on curve $B$ and ${n\choose k}$ is the number of $k$-combinations.
If we denote the components on $x,y,z$ axis of $B$ as $B^x,B^y,B^z$ and the components on $x,y,z$ axis of $P_k$ as $P_k^x,P_k^y,P_k^z$ respectively, \textit{e.g.}, $B=(B^x,B^y,B^z)$, $P_k=(P_k^x,P_k^y,P_k^z)$.
Then Eq.~(\ref{eq:bezier_curve}) can be rewritten as:

\begin{equation}
    \left\{
    \begin{array}{cc}
    B^x(\tau) &=\sum_{k=0}^{n}{n\choose k} \tau^{k}(1-\tau)^{n-k}P_k^x \\
    B^y(\tau) &=\sum_{k=0}^{n}{n\choose k} \tau^{k}(1-\tau)^{n-k}P_k^y \\
    B^z(\tau) &=\sum_{k=0}^{n}{n\choose k} \tau^{k}(1-\tau)^{n-k}P_k^z \\
    \end{array}
    \right.,
    \label{eq:bezier_curve_components}
\end{equation}

\noindent
similarly, we denote boundary $C=(C^x,C^y,C^z)$ where $C^x,C^y,C^z$ are the boundary $C$'s components on axis $x,y,z$.
We denote a point on the skeleton $C$ as $C(\tau)=(C^x(\tau),C^y(\tau),C^z(\tau))\ (\tau\in[0,1])$.
The optimization problem in Eq.~(\ref{eq:optimize}) as follows:

\begin{equation}
    {\rm min}||C-B||^2 \Leftrightarrow 
    \left\{
    \begin{array}{c}
        {\rm min}||C^x-B^x||^2 \\
        {\rm min}||C^y-B^y||^2 \\
        {\rm min}||C^z-B^z||^2 \\
    \end{array}
    \right..
\end{equation}

For simplicity, we will optimize ${\min}||C^x-B^x||^2$ as example and the optimization problem can be expanded as follows:

\begin{equation}
    \begin{array}{c}
    \min ||B^x-C^x||^2=\int_{0}^{1}||B^x(\tau)-C^x(\tau)||^2d\tau=\\
    \underset{{\rm card}(T)\to\infty}{\lim}\underset{\tau_i\in T}{\sum}||B^x(\tau_i)-C^x(\tau_i)||^2=\\
    \underset{{\rm card}(T)\to\infty}{\lim}\underset{\tau_i\in T}{\sum}||\sum_{k=0}^{n}{n\choose k}\tau_{i}^{k}(1-\tau_{i})^{n-k}P_k^x-C^x(\tau_i)||^2
    \end{array},
\end{equation}

\noindent
where $T=\{\tau_0,\tau_1,\cdots,\tau_m\}$ is a set of uniformly sampled points of $\tau\in[0,1]$ and ${\rm card}(T)=m+1$ is the cardinality of $T$.
If we denote $a_{ik}={n\choose k}\tau_{i}^{k}(1-\tau_{i})^{n-k}$. Thus, we have:

\begin{equation}
    \begin{array}{c}
    \underset{\tau_i\in T}{\sum}||\sum_{k=1}^{n}a_{ik}P_k^x-C^x(\tau_i)||^2=||\mathbf{Ap}-\mathbf{b}||^2=\\
    \left\Vert
    \left[\begin{array}{ccc}a_{00} & \cdots & a_{0n} \\ \vdots & \ddots & \vdots \\ a_{m0} & \cdots & a_{mn} \end{array}\right]
    \left[\begin{array}{c} P_0^x \\ \vdots \\ P_n^x \end{array}\right]-
    \left[\begin{array}{c} C^x(\tau_0) \\ \vdots \\ C^x(\tau_m) \end{array}\right]
    \right\Vert^2
    \end{array},
    \label{eq:least_square}
\end{equation}

\noindent
where $\mathbf{A}\in\mathbb{R}^{(m+1)\times (n+1)},\ \mathbf{p}\in\mathbb{R}^{(n+1)\times 1},\ \mathbf{b}\in\mathbb{R}^{(m+1)\times 1}$.
We need to solve the $\mathbf{p}$ by minimizing $||\mathbf{Ap}-\mathbf{b}||^2$.
Thus, the optimization problem in Eq.~(\ref{eq:original_problem}) is converted to solve the least square problem Eq.~\ref{eq:least_square}.
The solution $\hat{\mathbf{p}}={\rm arg\,min}_{\mathbf{p}}||\mathbf{Ap}-\mathbf{b}||^2$ can be computed by Gauss–Newton algorithm or $\hat{\mathbf{p}}=\mathbf{A}^{\dag}\mathbf{p}$ where $\mathbf{A}^{\dag}$ is the pseudo inverse matrix of $\mathbf{A}$.

\subsection{Motion Decay Along Curve Scale}

\begin{figure}[h]
    \centering
    \includegraphics[width=1.0\linewidth]{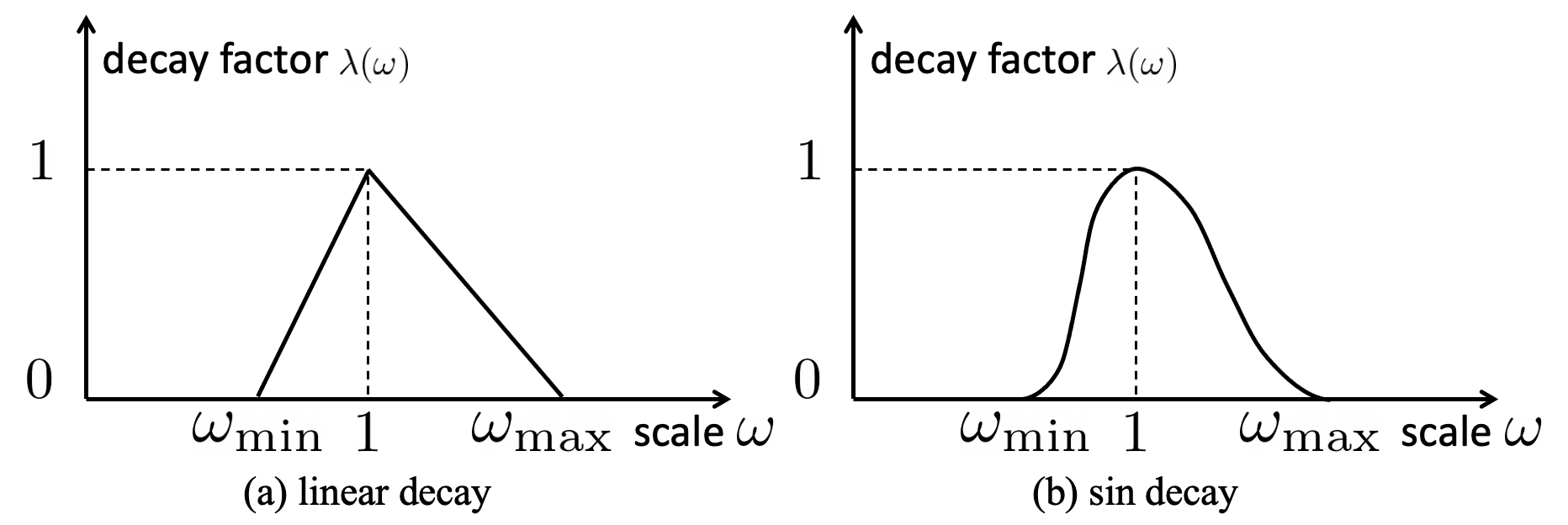}
    \caption{\textbf{Motion Decay.} (a) linear decay: the motion linearly decay with the curve scale when it deviate 1. (b) sin decay: we use the sine function to describe how the motion decays smoothly.}
    \label{fig:motion_decay}
\end{figure}

In the main paper, the motion at point $B_i(1,\tau_i)$ is denoted as $\mathcal{M}_{B_i(1,\tau_i)}^{e}$.
The point $B_i(1,\tau_i)$ lies at a composite Bézier curve $B_i$ that correspond to a motion seed.
The motion $\mathcal{M}_{B_i(1,\tau_i)}^{e}$ will decay when it spreads from $B_i(1,\tau_i)$ to $B_i(\omega_i,\tau_i)$. We have the following motion decay function:

\begin{equation}
    \mathcal{M}_{B_i(\omega_i,\tau_i)}^{e}=\lambda(\omega_i)\cdot\mathcal{M}_{B_i(1,\tau_i)}^{e},
\end{equation}

\noindent
where $\lambda(\omega_i)$ is the motion decay factor that is determined by the curve scale factor $\omega_i$.
In practice, we design two different decay functions as Fig.~\ref{fig:motion_decay} shows.
In case that the motion seed of the mouth spreads to the eyes area, we use $\omega_{\min}$ and $\omega_{\max}$ to restrict the area to where a motion seed can spread.
We find that performances of the \textit{linear decay} and \textit{sine decay} functions are similar.
Thus, we use the more simplified linear decay function in our experiments.
We leave the exploration of decay functions for the future work.

\subsection{Architecture of the Autoencoder $\mathcal{G}$}

The architecture of the Autoencoder network $\mathcal{G}$ is demonstrated in Tab.~\ref{tab:ae}. In the table, the \textbf{Resolution} denotes the spatial resolution of the feature map.
\textbf{EncBlock} denotes a 2D convolutional layer (stride is 2, padding is 1, kernel size is $4\times 4$).
\textbf{DecBlock} denotes a 2D convolutional layer (stride is 2, padding is 1, kernel size is $4\times 4$), followed by a PixelShuffle~\cite{shi2016real} layer (upscale factor is 2).

\begin{table}[h]
    \centering
    \caption{\textbf{Architecture of the AutoEncoder Network} $\mathcal{G}$}
    \resizebox{1\linewidth}{!}{
    \begin{tabular}{c|c|c}
    \Xhline{1pt}
    Layer Name & Resolution & Layer Structure \\
    \Xhline{1pt}
    Input & $256\times 256$ & Input Image \\
    $\mathcal{E}_1$ & $128\times 128$ & EncBlock $3\rightarrow 64$ + LeakyReLU(0.2) \\
    $\mathcal{E}_2$ & $64\times 64$ & EncBlock $64\rightarrow 128$ + LeakyReLU(0.2)\\
    $\mathcal{E}_3$ & $32\times 32$ & EncBlock $128\rightarrow 256$ + LeakyReLU(0.2)\\
    $\mathcal{D}_3$ & $64\times 64$ & DecBlock $256\rightarrow 128$ + ReLU\\
    $\mathcal{D}_2$ & $128\times 128$ & DecBlock $128\rightarrow 64$ + ReLU\\
    $\mathcal{D}_1$ & $256\times 256$ & DecBlock $64\rightarrow 3$\\
    \Xhline{1pt}
    \end{tabular}
    }
    \label{tab:ae}
\end{table}

\section{Implementation Details}
\label{sec:imp_details}

In the Parametric Shape Modeling, we find that it is hard to define nose, ears, eyebrows and jawline for pareidolia faces.
Thus, we only animate the mouth and eyes of pareidolia faces.
The mouth of a pareidolia face is animated by the inner lip boundary of the given human video.
For the mouth or eyes of a human/pareidolia face, its boundary is splitted into two branches (upper and lower halves) and each branch is parameterized as a composite Bézier curve.
In practice, we find that each branch of the mouth/eye can be precisely parameterized by a composite Bézier curve defined by 5-7 control points.

In the Expansionary Motion Transfer, for any pixel location $\mathbf{p}$, we compute $\omega_i,\tau_i$ that correspond to the composite Bézier curve $B_i$ (related to motion seed), which is prepared for our motion spread strategy.
Both the motion field and inverse motion field are computed on the discrete pixel grid.
We regard the motion field $\mathcal{M}^{e}$ as a function of the pixel grid and infer the inverse motion field $\overleftarrow{\mathcal{M}^{e}}$ as the inverse function of $\mathcal{M}^{e}$.
We use first-order difference of $\mathcal{M}^{e}$ in $\frac{d}{d\mathbf{p}}\overleftarrow{\mathcal{M}^{e}}(\mathbf{p})$.
In experiments, we find that the First-order Motion Approximation works well when $||\Delta\mathbf{p}||=1,2$ and increasing $||\Delta\mathbf{p}||$ does not bring further improvement.

In the Unsupervised Texture Animator, the image resolution is $256\times 256 $ for all the input human/pareidolia faces and the resolution of the output pareidolia faces is $256\times 256$, too.
During training the Autoencoder network $\mathcal{G}$, we set the loss weights $\alpha_1=\alpha_2=1$ empirically.

\section{Limitation}
\label{sec:limit}

\begin{figure}[h]
    \centering
    \includegraphics[width=0.95\linewidth]{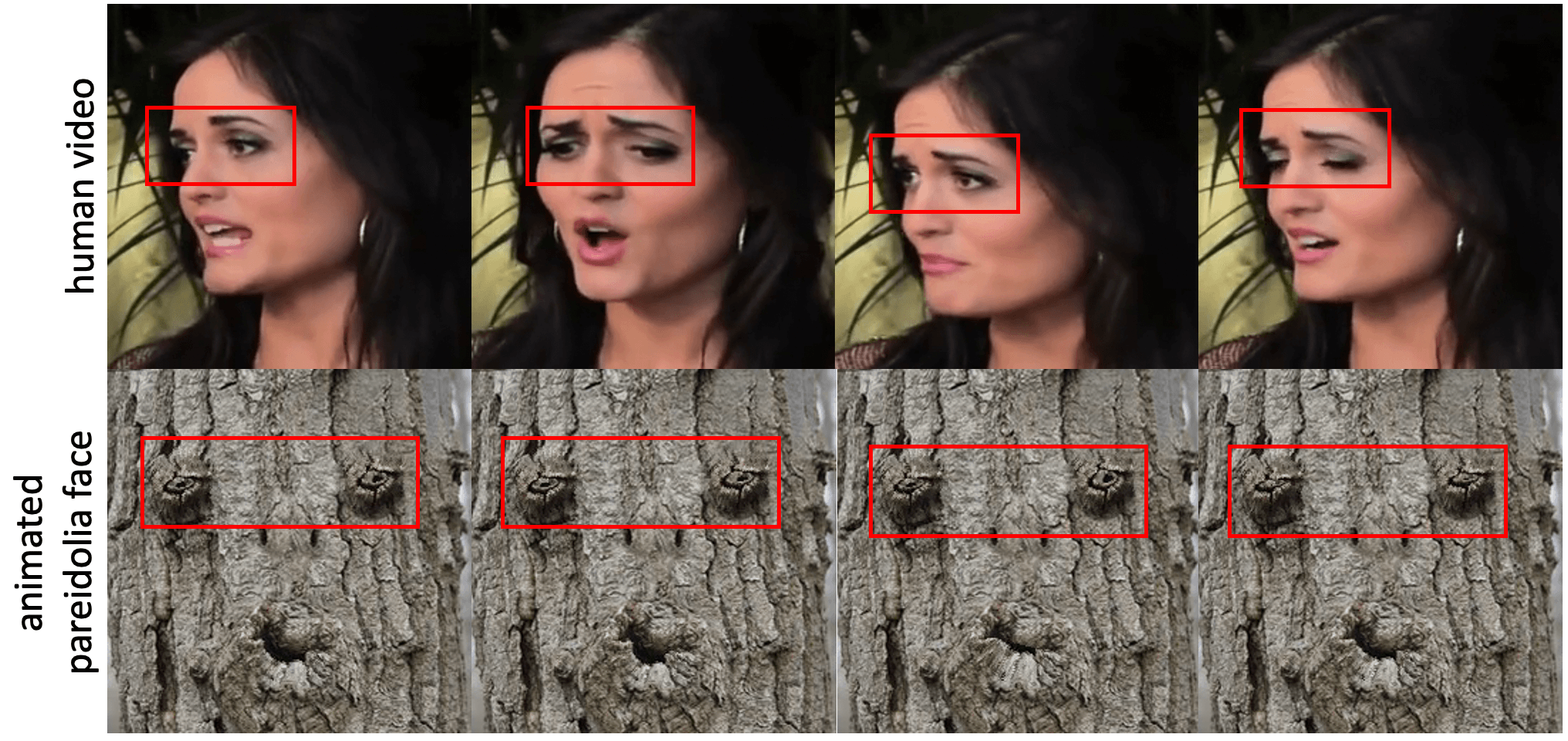}
    \caption{\textbf{A failure case caused by large poses of human faces.}
    Note that the pareidolia face does not well imitate the eyes movement of the human face.
    }
    \label{fig:large_poses}
\end{figure}

\textit{large poses of human faces:}
The facial motion extracted from human faces strongly relies on the robustness of the facial 3D landmark alignment tool.
For large poses of human faces, the 68 3D landmarks extracted by a face alignment tool might not be good enough, which makes the extracted facial motion of human faces inaccurate.
Then, the subsequent Expansionary Motion Transfer and Texture Animator will also be influenced.
We present some failed results in Fig.~\ref{fig:large_poses}.

\begin{figure}[h]
    \centering
    \includegraphics[width=0.95\linewidth]{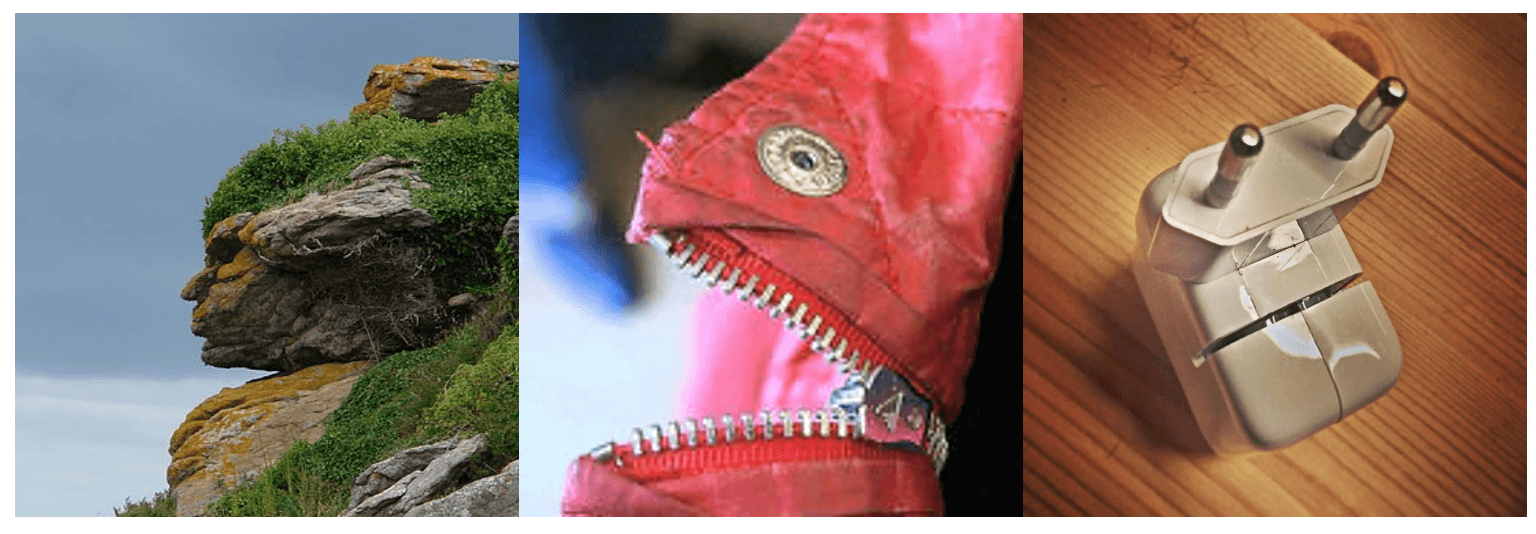}
    \caption{
    It is hard to label the facial boundaries for these pareidolia faces
    }
    \label{fig:boundary_extraction}
\end{figure}

\textit{automatic boundary extraction:}
Currently, our method requires us to label the facial boundaries for each input pareidolia face.
Thus, our pareidolia face reenactment is not a fully automatic method and we leave the automatic boundary extraction for pareidolia as future exploration.
In addition, as shown in Fig.~\ref{fig:boundary_extraction}, it is hard to label the facial boundaries for some pareidolia faces such as side faces. 

\begin{figure}[h]
    \centering
    \includegraphics[width=0.95\linewidth]{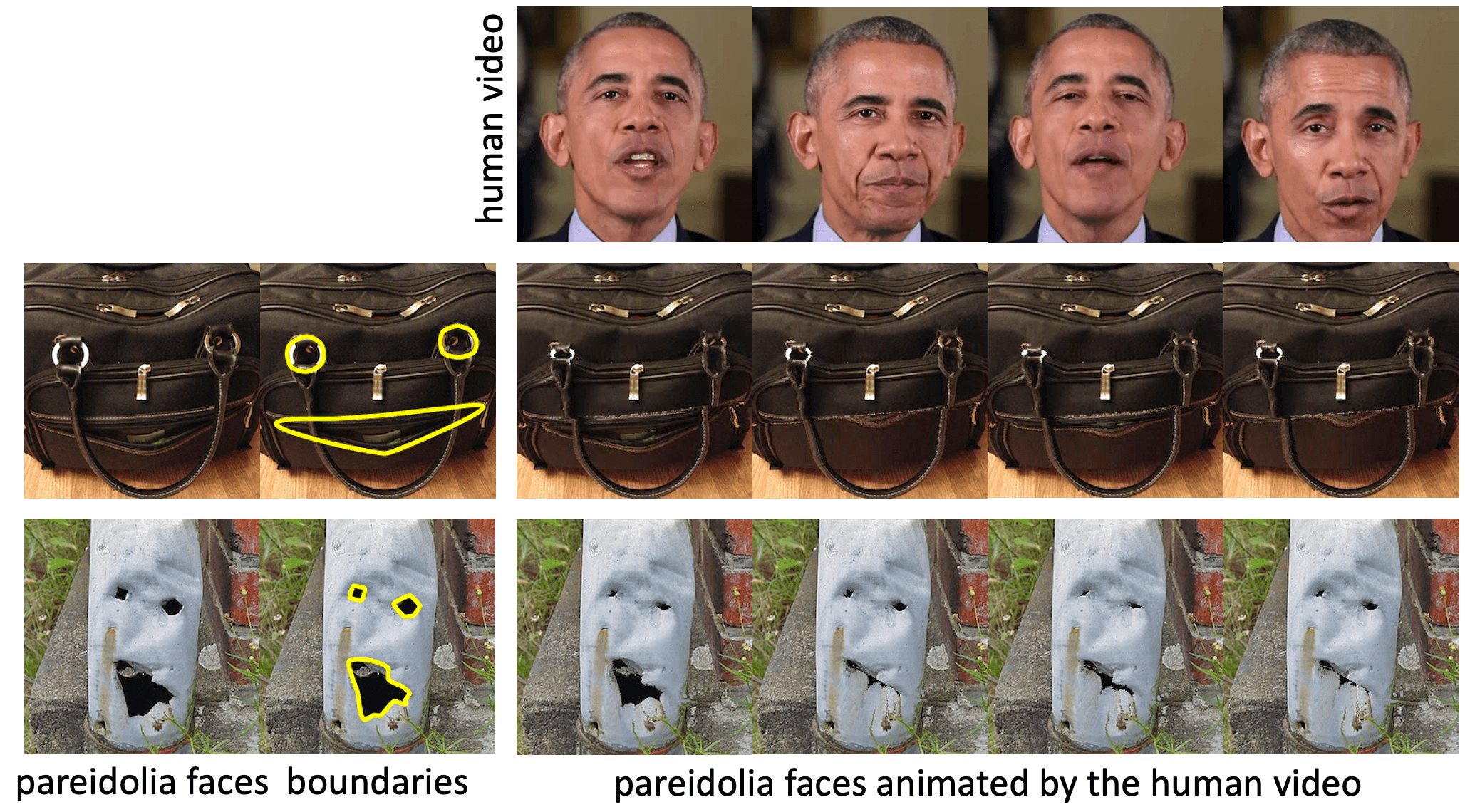}
    \caption{\textbf{Failure cases.}
    In the 2nd row, the texture of the bag's handle becomes incontinuity.
    In the 3rd row, the global structure of the mouth is broken.
    }
    \label{fig:failure}
\end{figure}

\textit{failure cases:}
For pareidolia faces with very complex boundaries and textures of facial parts, our proposed pareidolia face reenactment method might not well.
Note that we make the first attempt in animating a pareidolia face by the facial motion of a human face, we present some failure cases in Fig.~\ref{fig:failure}.

\end{document}